\documentclass[fleqn,10pt]{wlscirep}
\usepackage[utf8]{inputenc}
\usepackage[T1]{fontenc}
\usepackage{caption}
\usepackage{subcaption}
\usepackage{siunitx}
\usepackage{multirow}
\usepackage{amsmath}
\usepackage{array}
\usepackage{booktabs}
\usepackage{algorithm}
\usepackage{algpseudocode}
\usepackage{svg}

\usepackage[normalem]{ulem}

\def\forsubmission{1}

\ifx\forsubmission\undefined
\definecolor{YB}{RGB}{0,150,255}
\newcommand{\JY}[1]{\textcolor{blue}{JY: #1}}
\newcommand{\CF}[1]{\textcolor{red}{CF: #1}}
\newcommand{\YB}[1]{\textcolor{YB}{YB: #1}}
\newcommand{\KVDB}[1]{\textcolor{green}{KVDB: #1}}
\newcommand{\TODO}[1]{\textcolor{blue}{TODO: #1}}
\newcommand{\note}[1]{\textcolor{blue}{Note: #1}}
\newcommand{\red}[1]{\textcolor{red}{#1}}
\newcommand{\revision}[1]{\textcolor{black}{#1}}
\newcommand{\secondrev}[1]{\textcolor{red}{#1}}
\else 
\newcommand{\JY}[1]{\textcolor{blue}{}}
\newcommand{\CF}[1]{\textcolor{red}{}}
\newcommand{\YB}[1]{{}}
\newcommand{\KVDB}[1]{\textcolor{green}{}}
\newcommand{\TODO}[1]{\textcolor{red}{}}
\newcommand{\note}[1]{\textcolor{blue}{}}
\newcommand{\red}[1]{\textcolor{red}{}}
\newcommand{\revision}[1]{#1}
\newcommand{\secondrev}[1]{#1}
\fi

\def\thickhline{\noalign{\hrule height.8pt}}

\title{NeuroBench: A Framework for Benchmarking Neuromorphic Computing Algorithms and Systems}

\author[1,*]{Jason Yik}
\author[1,2]{Korneel Van den Berghe}
\author[2]{Douwe den Blanken}
\author[3]{Younes Bouhadjar}
\author[4]{Maxime Fabre}
\author[2,5]{Paul Hueber}
\author[6]{Weijie Ke}
\author[6]{Mina A Khoei}
\author[7,8]{Denis Kleyko}
\author[9]{Noah Pacik-Nelson}
% ...

\affil[1]{Harvard University}
\affil[2]{Delft University of Technology}
\affil[3]{Forschungszentrum Jülich}
\affil[4]{University of Groningen}
\affil[5]{imec}
\affil[6]{SynSense}
\affil[7]{Örebro University}
\affil[8]{RISE Research Institutes of Sweden}
\affil[9]{Accenture Labs}
% ...
\affil[*]{Correspondence to: jyik@g.harvard.edu}
% ...


\begin{abstract}
Neuromorphic computing shows promise for advancing computing efficiency and capabilities of AI applications using brain-inspired principles. However, the neuromorphic research field currently lacks standardized benchmarks, making it difficult to accurately measure technological advancements, compare performance with conventional methods, and identify promising future research directions. Prior neuromorphic computing benchmark efforts have not seen widespread adoption due to a lack of inclusive, actionable, and iterative benchmark design and guidelines. To address these shortcomings, we present NeuroBench: a benchmark framework for neuromorphic computing algorithms and systems. NeuroBench is a collaboratively-designed effort from an open community of researchers across industry and academia, aiming to provide a representative structure for standardizing the evaluation of neuromorphic approaches. The NeuroBench framework introduces a common set of tools and systematic methodology for inclusive benchmark measurement, delivering an objective reference framework for quantifying neuromorphic approaches in both hardware-independent (algorithm track) and hardware-dependent (system track) settings. In this article, we outline tasks and guidelines for benchmarks across multiple application domains, and present initial performance baselines across neuromorphic and conventional approaches for both benchmark tracks. NeuroBench is intended to continually expand its benchmarks and features to foster and track the progress made by the research community.

\end{abstract}

\begin{document}

\flushbottom


\maketitlewithpreprint

\thispagestyle{empty}



\section*{Introduction} 

In recent years, the rapid growth of artificial intelligence (AI) and machine learning (ML) has resulted in increasingly complex and large models in pursuit of higher accuracy and range of use cases~\cite{sevilla22computetrends}.
The substantial growth rate of model computation exceeds efficiency gains realized through Moore and Dennard technology scaling~\cite{shankar22trends}, indicating a looming limit to continued advancements with existing techniques.
This issue is compounded by the open challenges of adapting such methods for resource-constrained edge devices (tinyML) in order to enable pervasive and decentralized intelligence through the Internet of Things (IoT)~\cite{ray22tinyml}.
As such, the urgency for exploring new resource-efficient and scalable computing architectures has intensified.

Neuromorphic computing has emerged as a promising area in addressing these challenges, aiming to unlock key hallmarks of biological intelligence by porting primitives and computational strategies employed in the brain into engineered computing devices and algorithms~\cite{schuman2017survey, james17historical, thakur2018largescale}.
Neuromorphic systems hold a critical position in the investigation of novel architectures, as the brain exemplifies an exceptional model for accomplishing scalable, energy-efficient, and real-time embodied computation.

Initially, the term ``neuromorphic'' referred specifically to approaches that aimed to emulate the biophysics of the brain by leveraging physical properties of silicon, as proposed by Mead in the 1980's~\cite{Mead:1990hw}. However, the field of neuromorphic computing research has since grown to encompass a wide range of brain-inspired computing techniques at the algorithmic, hardware, and system levels~\cite{schuman2017survey}. 
While the range of approaches is diverse, neuromorphic computing research generally utilizes mechanisms emulating or simulating biophysical properties more closely than conventional methods, aiming to reproduce high-level performance and efficiency characteristics of biological neural systems. 

Neuromorphic \textit{algorithms}~\cite{Schuman2022} encompass neuroscience-inspired methods which strive towards goals of expanded learning capabilities, such as predictive intelligence, data efficiency, and adaptation, and include approaches such as spiking neural networks (SNNs) and primitives of neuron dynamics, plastic synapses, and heterogeneous network architectures.
Algorithm exploration often makes use of simulated execution on readily-available conventional hardware such as CPUs and GPUs, with the goal of driving design requirements for next-generation neuromorphic hardware. 

Neuromorphic \textit{systems}~\cite{frenkel23bottom} are composed of algorithms deployed to hardware, which seek greater energy efficiency, real-time processing capabilities, and resilience compared to conventional systems. Neuromorphic hardware utilizes a variety of biologically-inspired hardware approaches, including analog neuron emulation, event-based computation, non-von-Neumann architectures, and in-memory processing. Neuromorphic systems target a wide range of applications, from neuroscientific exploration, to low-power edge intelligence and datacenter-scale acceleration.

Despite its promises, progress in the field of neuromorphic research is impeded due to the absence of fair and widely-adopted objective metrics and benchmarks~\cite{Davies19_bencprog, Schuman2022}. Without such benchmarks, the validity of neuromorphic solutions cannot be directly quantified, hindering the research community from measuring technological advancement.
Standard and rigorous benchmarking is necessary for the neuromorphic community to objectively assess and compare the achievements of novel approaches, and make evidence-based decisions on which directions show promise for achieving breakthrough efficiency, speed, and intelligence, thereby helping to focus research and commercialization efforts on techniques that concretely improve on prior work and conventional computing. Neuromorphic benchmarks have been previously proposed for classical vision~\cite{orchard2015converting,Amir_etal17_lowpowe} and audition tasks~\cite{Cramer_2022}, open-loop~\cite{Ostrau2022} and closed-loop~\cite{closedloop-andre} tasks, and for SNN simulator performance assessment~\cite{KULKARNI2021}. While prior works have made valuable contributions, there are opportunities to further advance the field by addressing three outstanding challenges:

\begin{itemize}
    \item \textbf{Lack of a formal definition.} The variety of approaches to exploring brain-inspired principles creates difficulties in defining a set of criteria for what should be benchmarked as a ``neuromorphic” solution. Closed definitions can impose narrow assumptions and thus risk unfairly excluding promising methods.
    This challenge necessitates \textbf{\textit{inclusive}} benchmarks that can be applied generally across the spectrum of potential approaches, allowing for flexible implementation while focusing on task capabilities and metrics of interest such as temporal processing and efficiency.
    Furthermore, the benchmarks should ideally allow for direct comparison of neuromorphic and conventional approaches. 
    
    \item \textbf{Implementation diversity.} A wide array of different frameworks targeting different goals, such as neuroscientific exploration~\cite{nest2007} and automatic SNN training~\cite{eshraghian2021training}, are used in neuromorphic research. This diversity, which has been instrumental in exploring the landscape of bio-inspired techniques following different methodologies and abstraction levels, comes at the cost of portability and standardization, which in turn limits the ease of benchmark implementation. Benchmarks require common infrastructure that unites tooling to enable \textbf{\textit{actionable}} implementation and comparison of new methods.
    
    \item \textbf{Rapid research evolution.} Neuromorphic approaches are continually and rapidly evolving as part of an emerging field. As the research community continues to make technological progress, so too should benchmark suites and methodology expand to foster inclusion and capture salient performance metrics. An \textbf{\textit{iterative}} benchmark framework with structured versioning will facilitate productive foundational and evolving performance evaluation.
\end{itemize}

\begin{figure}[t]
\centering
\includegraphics[width=\linewidth]{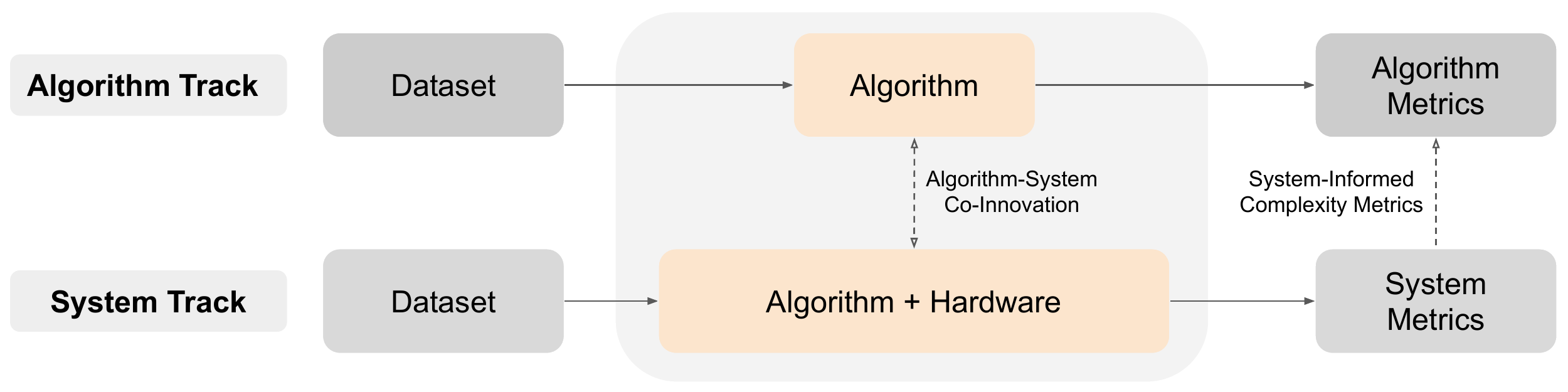}
\caption{The two NeuroBench tracks: algorithms and systems. Grey boxes designate what is defined by the benchmark, and orange boxes indicate what is unique to each solution. Connecting arrows between the two tracks denote the co-innovation between the tracks and the cross-stack innovation enabled by this approach. Between algorithm and system solutions, best-performing results from each track can motivate future solutions to the other. In addition, system metrics and results can inform hardware-independent algorithmic complexity metrics. 
}
\label{fig:tracks}
\end{figure}

To tackle these challenges, this article presents NeuroBench, a dual-track, multi-task benchmark framework. 
NeuroBench addresses the existing neuromorphic benchmark challenges by advancing prior work in three distinct ways. Firstly, the benchmark framework reduces assumptions regarding the specific solution being assessed, encouraging \textit{inclusive} participation of neuromorphic and non-neuromorphic approaches by utilizing general, task-level benchmarking and hierarchical metric definitions which capture key performance indicators of interest. Secondly, the NeuroBench benchmarks are associated with a common open-source benchmark harness tool which facilitates \textit{actionable} benchmark implementation and offers structure for further expansion to neuromorphic algorithm frameworks and systems. Finally, NeuroBench establishes an \textit{iterative}, community-driven initiative designed to evolve over time to ensure representation and relevance to neuromorphic research, analogous to the well-established MLPerf benchmark framework for machine learning~\cite{Reddi2020, mattson2020mlperf}.
As a whole, NeuroBench intends to align the neuromorphic research community on standard benchmarking, providing a dynamically evolving platform to ensure ongoing relevance and facilitate advancements through workshops, competitions, and a centralized leaderboard.

As Figure~\ref{fig:tracks} shows, the NeuroBench framework involves two tracks to enable agile algorithm and system development. As an emerging technology, neuromorphic hardware has not converged to a single platform which is commercially available, thus a large fraction of neuromorphic research explores algorithmic advancement on conventional systems which may not be optimal for performance. Thus, NeuroBench consists of an \textit{algorithm} track for hardware-independent evaluation and a \textit{system} track for fully deployed solutions. The algorithm track defines four novel benchmarks for neuromorphic methods across diverse domains, namely few-shot continual learning, computer vision, motor cortical decoding, and chaotic forecasting, and utilizes complexity metrics to analyze solution costs. Such hardware-independent benchmarking enables algorithmic exploration and prototyping, especially when simulating algorithm execution on non-neuromorphic platforms. Meanwhile, the system track defines standard protocols to measure the real-world speed and efficiency of neuromorphic hardware on benchmarks ranging from standard machine learning tasks to promising fields for neuromorphic systems, such as optimization. 

Each NeuroBench track includes defined datasets, metric and measurement methodology, and modular evaluation components to enable flexible development. Promising methods identified from the algorithm track will inform system design by highlighting target algorithms for optimization and relevant system workloads for benchmarking. The system track in turn enables optimization and evaluation of performant implementations, providing feedback to refine algorithmic complexity modeling and analysis. The interplay between the tracks creates a virtuous cycle: algorithm innovations guide system implementation, while system-level insights accelerate further algorithmic progress. This approach allows NeuroBench to advance neuromorphic algorithm-system co-design. Both the algorithm and system track will be extended and co-developed as NeuroBench continues to expand.

In the next few sections, we describe the algorithm track, including general complexity metric definitions, benchmark tasks, and common infrastructure tooling. We apply the framework to report baseline results for each algorithm benchmark, which outline unexplored research opportunities in optimizing algorithmic architectures and training of sparse, stateful models to achieve greater performance and resource efficiency. \secondrev{Then, we show baseline results established in the system track to assess neuromorphic performance across promising application workloads.}
By outlining both tracks, we provide a roadmap towards standardizing benchmark procedures in both hardware-independent and hardware dependent settings.


\section*{Algorithm Track Benchmark Framework}
\begin{figure}[t]
            \centering
            \includegraphics[width=.95\textwidth]{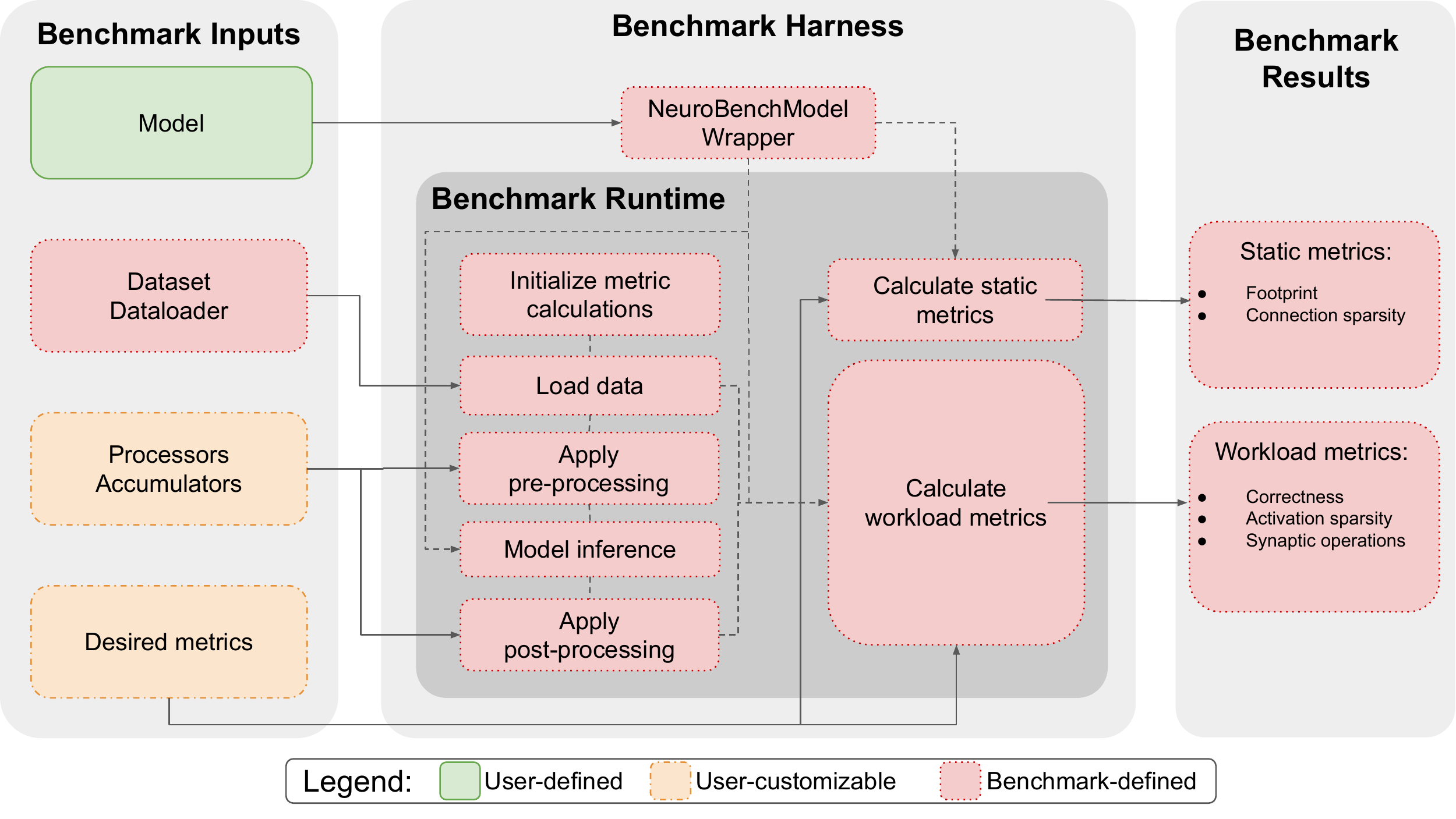}
            \caption{An overview of the NeuroBench algorithm track.}
        \label{fig:AlgTrack_SA}
    \end{figure}

The algorithm benchmark track aims to evaluate algorithms in a system-independent manner, separating algorithm performance from specific implementation details. The implementation platform can thus be ill-matched to the particular algorithm benchmark that it executes (e.g., SNN execution via dense matrix multiplication on a GPU), and the algorithm complexity and expected performance can be examined in a theoretical manner, motivating agile prototyping and functional analysis. Furthermore, minimal assumptions are made about the solutions tested, promoting inclusion of diverse algorithmic approaches.

The framework, as illustrated in Figure~\ref{fig:AlgTrack_SA}, is composed of inclusively-defined benchmark metrics, datasets and data loaders, and common harness infrastructure, shown in red. The metrics focus on assessing algorithm correctness on specific tasks as well as capturing general metrics that reflect the architectural complexity, computational demands, and storage requirements of the models. The datasets and data loaders specify the details of the tasks used for evaluation and ensure consistency across benchmarks. Finally, the harness infrastructure automates runtime execution and result output for the algorithm benchmark specified by the input interface, which consists of the user's model and customizable components for data processing and desired metrics, shown in green and orange.




    \subsection*{Algorithm Track Metrics}
    The algorithm track establishes \textit{solution-agnostic} primary metrics which are generally relevant to all types of solutions, including artificial and spiking neural networks (ANNs, SNNs). Firstly, there are \textit{correctness} metrics, which measure the quality of the model predictions on the particular task, such as accuracy, mean average precision (mAP), and mean-squared error (MSE). The correctness metrics are specified per task for each benchmark. Next, there are \textit{complexity} metrics, which measure the computational demands of the algorithm. In the first iteration of the NeuroBench algorithm track, we assume a digital, time-stepped execution of the algorithm and define the following complexity metrics:

    \begin{itemize}
        \item \textbf{Footprint} -- A measure of the memory footprint, in bytes, required to represent a model, which reflects quantization, parameters, and buffering requirements. The metric summarizes (and can be further broken down into) synaptic weight count, weight precision, trainable neuron parameters, data buffers, etc. Zero weights are included, as they are distinguished in the \textit{connection sparsity} metric. 

        
        \item \textbf{Connection Sparsity} -- 
        For a given model, the connection sparsity is the number of zero weights divided by the total number of weights, accumulated over all layers.
        0 refers to no sparsity (fully connected) and 1 refers to full sparsity (no connections). This metric accounts for deliberate pruning and sparse network architectures. 
        
        \item \textbf{Activation Sparsity} -- During execution, the average sparsity of neuron activations over all neurons in all model layers, for all timesteps of all tested samples, where 0 refers to no sparsity (i.e., all neurons are always activated), and 1 refers to the case where all neurons have a zero output.

        \item \textbf{Synaptic Operations} -- Average number of synaptic operations per model execution, based on neuron activations and the associated fanout synapses.
        This metric is further subdivided into dense, effective multiply-accumulate, and effective accumulate synaptic operations (Dense, Eff\_MACs, Eff\_ACs). Dense accounts for all zero and nonzero neuron activations and synaptic connections, and reflects the number of operations necessary on hardware that does not support sparsity. Eff\_MACs and Eff\_ACs only count effective synaptic operations by disregarding zero activations (e.g., produced by the ReLU function in an ANN or no spike in an SNN) and zero connections, thus reflecting operation cost on sparsity-aware hardware. Synaptic operations with non-binary activation are considered multiply-accumulates (MACs), while those with binary activation are considered accumulates (ACs).
    \end{itemize}

    Footprint and connection sparsity are classified as \textit{static metrics}, which can be analytically determined from the model only. Activation sparsity, synaptic operations, and correctness are classified as \textit{workload metrics}, which are dependent on execution or simulation of the model based on the benchmark data.

    \revision{In addition to the above complexity metrics, the algorithm track proposes to define \textit{Model Execution Rate}, corresponding to the rate, in Hz, at which the model's forward inference pass needs to be executed. For example, if a model is designed to process data from an event camera with a 50 ms input stride, the model execution rate is 20 Hz. The execution rate is a critical feature of the algorithm which provides intuition into the tradeoff between latency and computational footprint of a \textit{deployed} model, and is reported directly by the solution designer in benchmark results since it neither needs to be calculated nor extracted from the model or its outputs.}

    
    The complexity metrics are measured independently of the underlying hardware and therefore do not explicitly correlate with post-deployment latency or energy consumption. However, they provide valuable insight into algorithm performance and resource requirements, enabling high-level comparison and facilitating prototyping. For instance, the execution rate and number of synaptic operations can be taken together to estimate the speed and dynamic power of a model deployed to certain hardware, and the footprint and connection sparsity can be used to proxy hardware resource utilization. 

    Furthermore, the algorithm track can be extended with \textit{solution-specific} secondary metrics, which can offer deeper insights by using information specific to particular types of solutions. For example, for algorithms geared towards analog hardware, noise robustness is an important solution-specific metric. In addition, approaches with complex neuron dynamics may warrant measuring the overall complexity of a neuron update (i.e., type and counts of operations necessary to simulate the update), which can be combined with the total number of neuron updates in a model pass to calculate the cost of state updates. Such solution-specific metrics are expected to be community-driven and will be included in future NeuroBench algorithm track releases.

\subsection*{Algorithm Track Benchmarks}

    The v1.0 iteration of the NeuroBench algorithm track includes four benchmarks for neuromorphic computing research. The benchmarks were chosen by the NeuroBench community to capture key ongoing challenges for neuromorphic algorithm design. The list of tasks highlights features which are relevant to neuromorphic research interests: few-shot continual learning, object detection utilizing the high dynamic range and temporal resolution of event cameras, sensorimotor decoding based on cortical signals, and low-dimensional predictive modeling useful for prototyping resource-constrained networks that are suitable for small mixed-signal systems.
    Benchmark tasks are listed below and summarized in Table \ref{tab:alg_benchmarks}. Detailed specifications of benchmark tasks are provided in the Methods section. 
    
              

\begin{table}[t]
    \centering
    \renewcommand{\arraystretch}{1.3}
    \begin{tabular}{lllm{0.23\linewidth}}
        \thickhline
        Task & Dataset & Correctness metric & Task description  \\\hline
        Keyword FSCIL & MSWC~\cite{mazumder21mswc} & Accuracy & Few-shot, continual learning of keyword classes. \\\hline
        Event Camera Object Detection & Prophesee 1MP Automotive~\cite{Perot2020}  & COCO mAP & Detecting automotive objects from event camera video.  \\\hline
        NHP Motor Prediction & Primate Reaching~\cite{makin-dataset} & R$^2$ & Predicting fingertip velocity from cortical recordings.  \\\hline
        Chaotic Function Prediction & Mackey-Glass time series~\cite{mackey1977oscillation} & sMAPE & Autoregressive modeling of chaotic functions.  \\\thickhline
    \end{tabular}
    \caption{NeuroBench algorithm track v1.0 benchmarks.}
    \label{tab:alg_benchmarks}
\end{table}

    \begin{itemize}
        \item \textbf{Keyword Few-Shot Class-Incremental Learning (FSCIL)} --
        Learning new tasks from a small amount of experiences while retaining knowledge of prior tasks is a hallmark of biological intelligence and a long-standing goal of general AI~\cite{kudithipudi22biological}. It is especially a key challenge to endow edge devices with the ability to adapt to their environments and users. This benchmark thus evaluates the capacity of a model to successively incorporate new keywords over multiple sessions (class-incremental), with only a handful of samples from the new classes to train with (few-shot). The FSCIL task is a recently established benchmark in the computer vision domain~\cite{tao2020fscil}, but it has not yet been adapted to other data modalities. Aligning with a neuromorphic interest in temporal data modalities, this benchmark introduces a FSCIL task with streaming audio data using the large Multilingual Spoken Word Corpus (MSWC)~\cite{mazumder21mswc} keyword classification dataset. The task is designed to be approached in two phases: pre-training and incremental learning. First, for pre-training, a set of 100 words spanning 5 base languages (English, German, Catalan, French, Kinyarwanda) with 500 training samples each are made available to train an initial model. Next, for incremental learning, the model undergoes 10 successive sessions to learn words from 10 new languages (Persian, Spanish, Russian, Welsh, Italian, Basque, Polish, Esparanto, Portuguese, Dutch) in a few-shot learning scenario. Each incremental session adds 10 words of the corresponding session language with only 5 training samples available per word. After each session, the model is tested in classification accuracy on all prior learned classes, including the 100 base pre-training classes and the few-shot-learned classes,
        therefore evaluating the FSCIL solution on its ability to learn new classes while retaining knowledge about the previously learned ones. Each session learns a new language, for a total knowledge base of 200 keywords by the end of the benchmark.

        \item \textbf{Event Camera Object Detection} --
        Object detection is a widely-used computer vision task with applications in robotics, autonomous driving, and surveillance. Such scenarios at the edge may require high energy efficiency and real-time performance, which can be achieved via event-based vision sensors~\cite{Gallego2022}. The event camera object detection benchmark uses the Prophesee 1 Megapixel automotive detection dataset~\cite{Perot2020}, a large labeled object detection dataset with over 15 hours of event camera video from the front of a car driving in various scenarios. Predetermined training, validation, and testing splits include \SI{11.2}{\hour}, \SI{2.2}{\hour}, and \SI{2.2}{\hour} of recording, respectively. Pedestrian, two-wheeler, and car object classes are used in evaluation, and correctness is measured using COCO mean average precision (mAP)~\cite{LinCOCO_2014}.
        
        \item \textbf{Non-human Primate (NHP) Motor Prediction} --
        Studying models which can accurately replicate features of biological computation presents opportunities in understanding sensorimotor behavior and developing closed-loop methods for future robotic agents. It also is foundational to the development of wearable or implantable neuro-prosthetic devices that can accurately generate motor activity from neural or muscle signals. This benchmark utilizes a dataset consisting of multi-channel recordings from the sensorimotor cortex of two non-human primates (NHP Indy and NHP Loco) during reaching movements, along with corresponding fingertip motion of the reach~\cite{makin-dataset}. Six total sessions are included from the dataset, for a total of 8712 seconds of data. The task is to train a model to predict the two-dimensional components of finger velocity using recent neural data. The sessions are treated independently (i.e., models are trained separately for each session), and the data is split to allow the first 75\% for training and validation and the last 25\% for evaluation. Correctness of the predictions is evaluated by the coefficient of determination ($R^{2}$) score against the true finger velocity targets, averaged over all six sessions.
        
        \item \textbf{Chaotic Function Prediction} --
        The real-world data benchmarks presented thus far are high-dimensional and can require large networks to achieve high accuracy, raising challenges for solution types with limited I/O support and network capacity, such as mixed-signal edge prototype solutions. To address this, we include a synthetic benchmark based on prediction of one-dimensional Mackey-Glass time series~\cite{mackey1977oscillation}, which can be effectively tackled by smaller networks. Mackey-Glass has been widely adopted as a benchmark for evaluating temporal predictors, including neuromorphic models~\cite{jaeger2004harnessing, mukhopadhyay2020learning, chilkuri2021parallelizing}. The task involves prediction of the next timestep value $f(t+\Delta t)$ given the current timestep value $f(t)$. 
        The model is trained and validated using the first half of the time series, during which the ground truth state $f(t)$ are supplied to the model to predict the next timestep $f'(t+\Delta t)$.
        During the evaluation, the model uses its prior prediction $f'(t)$ to generate each next value $f'(t+\Delta t)$, autoregressively forecasting the second half of the time series.
        Correctness is measured using symmetric mean absolute percentage error (sMAPE) of the generated time series against the target time series, a standard metric in forecasting~\cite{MAKRIDAKIS202054}.
        The benchmark includes a set of 14 Mackey-Glass time series, which vary by the equation parameter $\tau$, the delay constant. Lyapunov time $(L)$, the expected predictability timescale for chaos~\cite{gilpin2023largescale}, is used as the time unit for each time series. The total length of each series is 20 Lyapunov times, and 75 points are sampled per Lyapunov time ($\Delta t$ = $L/75$).

    \end{itemize}

\subsection*{Algorithm Track Benchmark Harness}
    The NeuroBench algorithm benchmarks are wrapped in a harness which standardizes the benchmark interfaces. The harness provides benchmark users with a consistent framework for loading data, processing data and model outputs, and calculating and reporting metrics, thereby ensuring fair and standard comparisons of the results. It is built with straightforward interfaces which are designed to be extended with new frameworks, algorithms, and tasks. The benchmark harness is open-source for use and development (\url{https://github.com/NeuroBench/neurobench}).
    
    The components of the algorithm benchmark harness are summarized in Figure~\ref{fig:AlgTrack_SA}.
    \textit{Datasets} are loaded in a common format and pass through \textit{Processors} to be pre-processed. The \textit{Model} generates predictions based on the processed data, and \textit{Accumulators} post-process the predictions, for instance to accumulate spikes and transform to labels. Static metrics of algorithm footprint and connection sparsity are calculated via model analysis, while metrics of correctness, activation sparsity, and synaptic operations are calculated using predictions and model execution traces. For benchmark users, task evaluation simply involves utilizing the existing dataloaders, processors, and metrics within the harness and wrapping their own code to fit the standard interfaces.
 
    Currently, the harness and all baseline models are built using PyTorch~\cite{paszke19pytorch} or frameworks based on it, such as snnTorch~\cite{eshraghian2021training} and SpikingJelly~\cite{SpikingJelly}. Due to its modular structure and simple interfaces, the harness can grow to be compatible with further neuromorphic tools such as Lava~\cite{lava} and Fugu~\cite{aimone2019composing}.
    Furthermore, it also supports the extension of data and metric pipelines in order to implement additional benchmark tasks. \revision{Widely validated benchmarks in keyword~\cite{warden2018speech} and gesture classification~\cite{Amir_etal17_lowpowe}, which are foundational in neuromorphic and conventional research~\cite{blouw2019, Cramer_2022, fang2021incorporating, massa2021efficient}, have been incorporated into the harness to complement the novel tasks in the NeuroBench v1.0 suite.} Any novel or existing benchmarks can make use of the harness infrastructure for open reproducibility, and also to garner interest in the community towards long-term task support and appearance in NeuroBench-affiliated leaderboards and challenge events.

\subsubsection*{Algorithm Track Limitations and Further Extensions}

Before diving into the baseline results, it is worth discussing several possible improvements to the NeuroBench algorithm track framework in its current form. Specifically, the initial iteration of metrics is restricted to the assumption of digital, time-stepped algorithm execution. While complexity analysis of such prototypes can serve as an intermediate step for solutions intended for analog or continuous time deployment, the metric measurements are not yet defined for those execution settings. Informed by further benchmark implementations, future versions of NeuroBench will extend inclusiveness by expanding measurement protocols to include such algorithms.

\revision{Furthermore, the synaptic operations metric, intended to capture model computation cost, currently does not account for neuron updates. The dynamics of neuron models, including mechanisms like leakage and reset, can vary heavily in complexity. However, counting the number and type of operations from neuron updates, as well as estimating their overall costs, depends on the specific arithmetic or circuit implementation. Thus, they are not accounted for in the broader algorithmic complexity metrics. The algorithmic metric framework can be extended with solution-specific metrics that assume a particular implementation platform to estimate neuron update costs, which have been previously defined~\cite{Lemaire_2023}. These estimates can then be combined with the total number of neuron updates per model computation to measure overall network operation complexity during evaluation.}


Data pre- and post-processing can also amount to significant costs not yet captured in the NeuroBench algorithm track metrics. Such costs are, however, captured in the deployed metrics of the system track, which accounts for data processing hardware as part of the overall system during performance and efficiency measurements. Data processing metrics will be added as a separate complexity category for the algorithm track benchmark in the future.

The v1.0 algorithm track benchmark suite is also intended to expand in the future. This could include covering further data modalities such as inertial measurement unit (IMU) sensing~\cite{fra22har} and extending to closed-loop sensorimotor tasks to demonstrate embodied intelligence. As with the initial benchmarks, further tasks will undergo approval and development by the open NeuroBench community before being included in a future versioned benchmark suite.

\subsection*{Algorithm Track Baseline Results}



In our first iteration of the algorithmic track, we report baseline algorithm performance on each benchmark using various model architectures, including artificial neural networks commonly used in deep learning, spiking neural networks, and reservoir networks. We evaluate each benchmark with two substantially different algorithm baselines. From these evaluations, we extract baseline comparisons, identify trends, and uncover motivations for future research. Except for the event camera object detection task, each benchmark utilizes a novel data split, and all tasks use novel metric measurement. The presented baselines are a snapshot of the solution search space and will be starting points for leaderboards, thereby calling for further research to push the state of the art for each task. Detailed specifications of each of the baselines can be found in the Methods section.

    \subsubsection*{Keyword FSCIL}

\begin{table}[ht]
\centerline{
\renewcommand{\arraystretch}{1.3}
\begin{tabular}{c|cccccccc} \thickhline
\multirow{2}{*}{Baseline} &

  \multirow{2}{*}{\begin{tabular}[c]{@{}c@{}}Accuracy\\ (Base / Session Avg)\end{tabular}} &
  \multirow{2}{*}{\begin{tabular}[c]{@{}c@{}}Footprint\\ (bytes)\end{tabular}} &
  \multirow{2}{*}{\begin{tabular}[c]{@{}c@{}}Model Exec.\\ Rate (Hz)\end{tabular}} &
  \multirow{2}{*}{\begin{tabular}[c]{@{}c@{}}Connection\\ Sparsity\end{tabular}} &
  \multirow{2}{*}{\begin{tabular}[c]{@{}c@{}}Activation\\ Sparsity\end{tabular}} &
  \multicolumn{3}{c}{SynOps (per model exec.)} \\ \cline{7-9} 
        &       &          &    &     &       & \multicolumn{1}{c}{Dense}    & \multicolumn{1}{c}{Eff\_MACs} & Eff\_ACs \\ \hline
M5 ANN & (97.09\% / 89.27\%) & $6.03 \times 10^6$ & 1 & 0.0 & 0.783 & \multicolumn{1}{c}{$2.59 \times 10^7$} & \multicolumn{1}{c}{$7.85 \times 10^6$}  & 0 \\ 
SNN  & (93.48\% / 75.27\%) & $1.36 \times 10^7$ & 200 & 0.0 & 0.916 & \multicolumn{1}{c}{$3.39 \times 10^6$} & \multicolumn{1}{c}{0}  & $3.65 \times 10^5$ \\ \thickhline
\end{tabular}
}
\caption{Baseline results for the keyword few-shot class-incremental learning task. Base accuracy refers to accuracy on the 100 base classes after pre-training while session average accuracy is the average accuracy over all sessions for the corresponding prototypical baseline. The detailed accuracy per session for the different baselines are shown in Figure~\ref{fig:mswc_acc_per_session}.}
    \label{tab:FSCIL_results}
\end{table}



    The keyword FSCIL task has an ANN and SNN baseline, using different model architectures:
    
    \begin{itemize}
        \item \textbf{M5 ANN} -- The ANN baseline uses a tuned version of the M5 deep convolutional network architecture~\cite{dai2016deep}, with samples pre-processed into Mel-frequency cepstral coefficients (MFCC). The network contains four successive convolution-normalization-pooling layers, followed by a readout fully-connected layer. Each model execution (forward pass) uses the data from the full pre-processed sample, and convolution kernels are applied over the temporal dimension of the samples. This is reported as a 1~Hz model execution rate.

        \item \textbf{SNN} -- The SNN baseline uses a recurrent SNN with adaptive leaky integrate-and-fire (LIF) neurons and heterogeneous time constants~\cite{bittar22surrogate}. The SNN consists of two recurrent adaptive LIF layers and one linear output layer.
        Audio samples are pre-processed to binary spike trains using Speech2Spikes~\cite{stewart23speech2spikes}, which relies on a Mel Spectrogram with the same parameters as the MFCC of the ANN baseline. Each input timestep to the model represents 5~ms of audio data, thus the model has a 200~Hz model execution rate. Output neuron activations are summed over time to produce the word class prediction.
    \end{itemize}



    After pre-training using standard batched training, the ANN and SNN baseline networks reach high accuracies on the base classes of 97.09\% and 93.48\%, respectively.
    As reported by the model execution rate metric, the SNN baseline computes each sample over 200 passes, using an order of magnitude fewer effective AC synaptic operations compared to the ANN baseline's effective MACs per model execution.
    Considering both the model execution rate and synaptic operation metrics, the number of aggregated ACs over the length of the sample ($200 * 3.65 \times 10^5 = 7.30 \times 10^7$) exceeds the Dense and effective MAC operations necessary for the ANN baseline, which spatially flattens the sample and processes it in one model execution.
    However, outside of the static-length keyword classification scenario, the low-cost per-execution temporal processing of SNNs can enable efficient, always-on, high-frequency prediction capabilities in deployed continuous audio recognition scenarios. 


    \begin{figure}[ht]
                \centering
                \includegraphics[width=\textwidth]{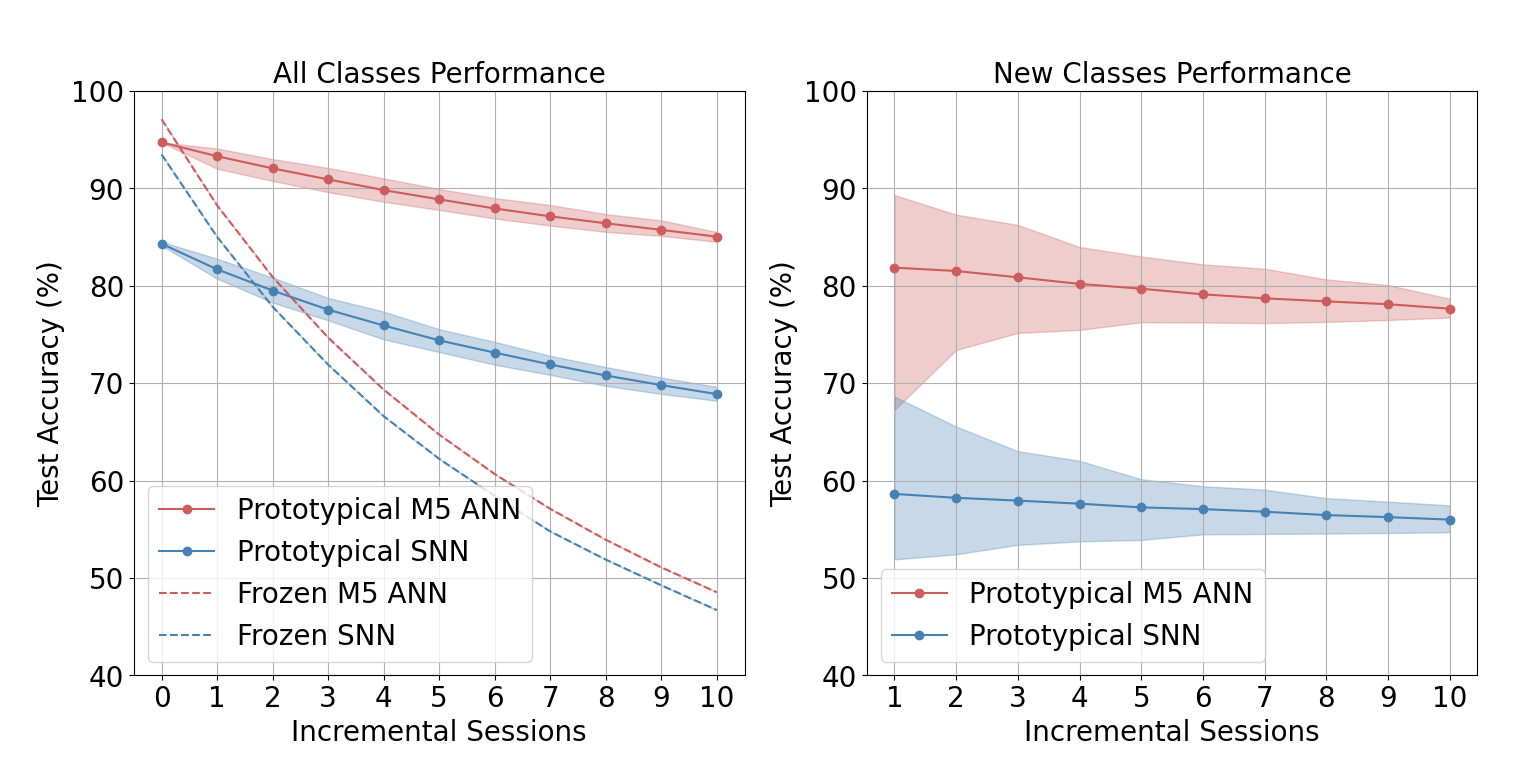}
                \caption{Test accuracy per session on the keyword FSCIL task for prototypical and frozen baselines, with the accuracy on both base classes and incrementally-learned classes (left), and accuracy on all incrementally-learned classes only (right). Incremental session 0 refers to the accuracy on base classes after pre-training only. Shaded area represents 5$^{th}$ and 95$^{th}$ percentile on 100 runs. Frozen baselines with no adaptation do not learn incremental classes and thus have a fixed 0\% accuracy for New Classes Performance. }
            \label{fig:mswc_acc_per_session}
        \end{figure}

    We present two approaches for the incremental stage for both the ANN and SNN baselines. The \textit{frozen} models are locked after pre-training on base classes and have 0\% accuracy on all new incremental classes, providing a reference for models with no learning or catastrophic forgetting of prior classes. The \textit{prototypical} models employ a prototypical network~\cite{snell2017prototypical} for incremental learning, which is a feature-based clustering approach that can be implemented with a simple linear readout layer on top of the pre-trained network backbone. Prototypical weights and biases of prior and incremental classes are directly defined based on the average features of the corresponding class and directly substitute pre-trained readout layer parameters.  The complexity results in Table~\ref{tab:FSCIL_results} thus empirically apply to both the frozen and prototypical models.
    
    The test accuracy for the baseline models over all sessions, as well as the test accuracy on only the new incrementally-learned classes, are shown in Figure~\ref{fig:mswc_acc_per_session}.
    Using prototypical networks, the ANN model reaches 89.27\% accuracy on average over all sessions, demonstrating significant greater performance of 21.41 accuracy points with respect to the frozen model. The accuracy on new classes, averaged over all incremental sessions, is 79.61\%.
    The SNN prototypical baseline, on the other hand, reaches 75.27\% accuracy on average over all sessions, surpassing the frozen SNN performance by 9.97 accuracy points, with an average accuracy on new classes over all sessions of 57.23\%. 

    The accuracy loss over the incremental sessions is similar between the ANN and SNN prototypical baselines. However, the lower overall accuracy of the SNN is largely due to the conversion from the original backpropagation-trained readout classifier, which is used in the frozen baseline, to the prototype readout classifier. 
    On the base classes (session 0 in Figure~\ref{fig:mswc_acc_per_session}), the ANN sees a drop of 2.37\% between the frozen and prototypical baselines, while the SNN has a larger drop of 9.17\%. 
    The larger drop indicates that our particular SNN baseline has a less general feature extraction than the ANN. This may be due to the challenges of backpropagation through time for online temporal inference to learn to extract long-term temporal keyword features with the chosen spiking recurrent model. Additionally, the Speech2Spikes~\cite{stewart23speech2spikes} pre-processing algorithm converting audio to spikes may also cause information loss. Overall, the keyword FSCIL benchmark presents opportunities for further research in learning methods, preprocessing, and model architectures for continual learning of temporal data.

    \subsubsection*{Event Camera Object Detection}


\begin{table}[ht]
\centerline{
\renewcommand{\arraystretch}{1.3}
\begin{tabular}{c|cccccccc} \thickhline
\multirow{2}{*}{Baseline} &
  \multirow{2}{*}{\begin{tabular}[c]{@{}c@{}}mAP\end{tabular}} &
  \multirow{2}{*}{\begin{tabular}[c]{@{}c@{}}Footprint\\ (bytes)\end{tabular}} &
  \multirow{2}{*}{\begin{tabular}[c]{@{}c@{}}Model Exec.\\ Rate (Hz)\end{tabular}} &
  \multirow{2}{*}{\begin{tabular}[c]{@{}c@{}}Connection\\ Sparsity\end{tabular}} &
  \multirow{2}{*}{\begin{tabular}[c]{@{}c@{}}Activation\\ Sparsity\end{tabular}} &
  \multicolumn{3}{c}{SynOps (per model exec.)} \\ \cline{7-9} 
        &       &          &    &     &       & \multicolumn{1}{c}{Dense}    & \multicolumn{1}{c}{Eff\_MACs} & Eff\_ACs \\ \hline
RED ANN & 0.429 & $9.13 \times 10^7$ & 20 & 0.0 & 0.634 & \multicolumn{1}{c}{$2.84 \times 10^{11}$} & \multicolumn{1}{c}{$2.48 \times 10^{11}$}  & 0 \\
Hybrid  & 0.271 & $1.21 \times 10^7$ & 20 & 0.0 & 0.613 & \multicolumn{1}{c}{$9.85 \times 10^{10}$} & \multicolumn{1}{c}{$3.76\times 10^{10}$}  & $5.60\times10^8$ \\ \thickhline
\end{tabular}
}
\caption{Baseline results for the event camera object detection task.}
    \label{tab:ObjDet_resuls}
\end{table}

    The event camera object detection task reports a prior baseline, the RED ANN, and a novel conversion of the architecture to a hybrid ANN-SNN model:
    
    \begin{itemize}
        \item \textbf{RED ANN} -- The RED architecture \cite{Perot2020} consists of blocks of feed-forward squeeze-and-excite~\cite{hu18squeeze} convolutional layers followed by blocks of recurrent convolution-LSTM (ConvLSTM~\cite{shi15convlstm}) layers.
        A single-shot detection (SSD~\cite{Liu_2016}) head is used to predict the location and class of the bounding box based on multi-scale outputs from the recurrent layers. 
        Raw event data is binned into 50 ms and pre-processed into time surfaces. 
        \item \textbf{Hybrid} -- The hybrid ANN-SNN architecture adopts feedforward LIF spiking neural layers to replace the ConvLSTM layers in RED, and shares the same feed-forward convolutional blocks as the RED. It uses the same input encoding method and SSD head as the RED model. 
    \end{itemize}

    Results for the two networks can be found in Table \ref{tab:ObjDet_resuls}. The RED ANN represents the current state-of-the-art correctness on the benchmark, at 0.429 mAP. The Hybrid network is a smaller network, reflected by the footprint and synaptic operations metrics measuring an order of magnitude smaller than for the RED ANN. The smaller size comes at the expense of lower correctness of 0.271 mAP.

    For the RED ANN, the activation sparsity metric (0.634) represents zero activations by the ReLU function for each neuron. From this, one may expect that the number of effective operations (operations with a nonzero activation and nonzero weight) would be around 35\% of dense operations, however the actual ratio is 87\%. This is due to the presence of normalization layers applied to activations before synaptic weight multiplication. Furthermore, neurons with lower activation frequency in the network tend to have a smaller fanout than neurons with high activation frequency. Thus, while activation sparsity alone can provide a proxy for the cost of the network, architectural characteristics may impede actual computation reduction, and the synaptic operations must be considered in tandem.

    The Hybrid network demonstrates a significant reduction in total effective operations against dense operations, outlining significant gains if deployed on specialized sparsity-aware hardware. However, for the particular network, the number of effective ACs, generated by the spiking neuron components, is two orders of magnitude smaller than the number of effective MACs within the ANN components. Such a hybrid network may not warrant specialized accumulation units, and the baseline motivates further research in hybrid networks with a larger proportion of spiking neuron activity compared to artificial neuron activity.


    \subsubsection*{NHP Motor Prediction}
\begin{table}[ht]
\centerline{
\renewcommand{\arraystretch}{1.3}
\begin{tabular}{c|cccccccc} \thickhline
\multirow{2}{*}{Baseline} &
  \multirow{2}{*}{\begin{tabular}[c]{@{}c@{}}$R^2$\end{tabular}} &
  \multirow{2}{*}{\begin{tabular}[c]{@{}c@{}}Footprint\\ (bytes)\end{tabular}} &
  \multirow{2}{*}{\begin{tabular}[c]{@{}c@{}}Model Exec.\\ Rate (Hz)\end{tabular}} &
  \multirow{2}{*}{\begin{tabular}[c]{@{}c@{}}Connection\\ Sparsity\end{tabular}} &
  \multirow{2}{*}{\begin{tabular}[c]{@{}c@{}}Activation\\ Sparsity\end{tabular}} &
  \multicolumn{3}{c}{SynOps (per model exec.)} \\ \cline{7-9} 
        &       &          &    &     &       & \multicolumn{1}{c}{Dense}    & \multicolumn{1}{c}{Eff\_MACs} & Eff\_ACs \\ \hline
\multirow{2}{*}{ANN}
 & 0.593  & 20824  & 250  & 0.0  & 0.683  & \multicolumn{1}{c}{4704}      & \multicolumn{1}{c}{3836}   & \multicolumn{1}{c}{0} \\ 
 & 0.558  & 33496  & 250  & 0.0  & 0.668  & \multicolumn{1}{c}{7776}      & \multicolumn{1}{c}{6103}  & \multicolumn{1}{c}{0} \\ \hline
\multirow{2}{*}{SNN}
 & 0.593  & 19648 & 250  & 0.0  & 0.997  & \multicolumn{1}{c}{4900}       & \multicolumn{1}{c}{0}    & \multicolumn{1}{c}{276} \\ 
 & 0.568 & 38848 & 250  & 0.0  & 0.999 & \multicolumn{1}{c}{9700}       & \multicolumn{1}{c}{0}    & \multicolumn{1}{c}{551} \\ \thickhline
\end{tabular}
}
\caption{Baseline results for the NHP motor prediction task, for NHP Indy (96-channel data, top), and NHP Loco (192-channel data, bottom).}
\label{tab:primate_results}
\end{table}

    Small fully-connected, feedforward networks were developed for the NHP motor prediction baselines:
    
    \begin{itemize}
        \item \textbf{ANN} -- In the ANN baseline, the cortical activity from the 50 most recent data samples is buffered to be used as network input. The network has two hidden layers and $2$ final outputs predicting $X$ and $Y$ velocities, with a fully-connected topology of $N_{ch}$-32-48-2, where $N_{ch}$ refers to the channels of cortical data (96 for NHP Indy, and 192 for NHP Loco). Batch normalization is applied after each hidden layer.
        
        \item \textbf{SNN} -- The SNN uses the data samples directly as input to the network, without buffering. It has a hidden layer of 50 LIF neurons, for a fully connected topology of $N_{ch}$-50-2 LIF neurons. The output neurons do not have a reset mechanism, and the membrane potential is directly read to produce the output velocities. 
        
    \end{itemize}

Table \ref{tab:primate_results} shows the results for the ANN and SNN baselines, averaged between sessions from each NHP (Indy and Loco). The ANN and SNN are similar in footprint size and number of dense operations per model forward pass, and also reach comparable prediction quality based on $R^2$ score. Each model is small in footprint and operation count, demonstrating that this task can be solved by shallow edge networks, validating prior studies~\cite{willsey22realtime}.

Between the baselines, the SNN realizes similar correctness at significantly reduced complexity compared to the ANN. Extremely high activation sparsity in the SNN (0.998) directly translates to low effective accumulate operations, demonstrating the adequacy of stateful, binary-activation neuron models for sparse regression tasks. Meanwhile, similarly to the RED ANN in the event camera object detection task, activation sparsity in the ANN baseline does not translate to effective operation efficiency, as batch normalization is applied to activations before multiplication with synaptic weights.

\begin{figure} [ht]
\centering
\begin{subfigure}{.5\textwidth}
  \centering
  \includegraphics[width=1\linewidth]{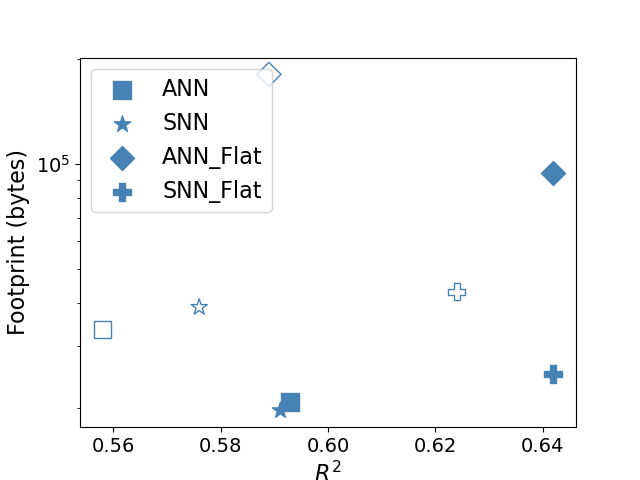}
  \label{fig:sub1}
\end{subfigure}%
\begin{subfigure}{0.5\textwidth}
  \centering
  \includegraphics[width=1\linewidth]{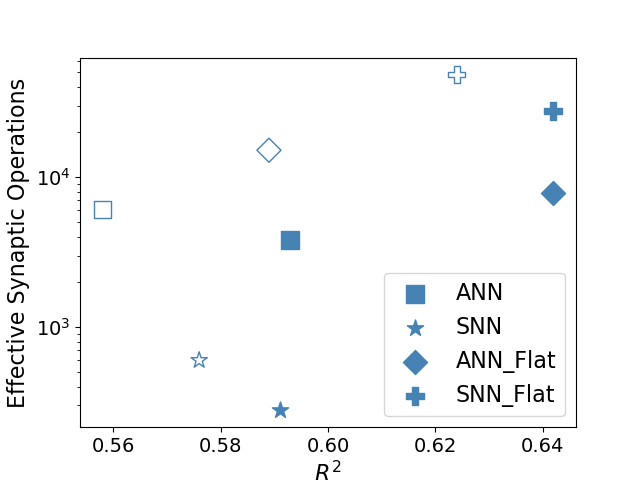}
  \label{fig:sub2}
\end{subfigure}
\caption{Footprint and effective synaptic operations vs $R^2$, for four task baselines. Each model has two points: the solid marker represents NHP Indy, and the hollow marker represents NHP Loco.}
\label{fig:motor_result}
\end{figure}

We conduct further exploration for increasing task accuracies with more complex ANN and SNN models: ANN\_Flat and SNN\_Flat. For these networks, 50 data samples of buffered input are split into $n_p=7$ accumulated bins. For ANN\_Flat, the 7 bins are spatially flattened as input to the network, so its topology is ($7\times N_{ch}$)-32-48-2. 
SNN\_Flat uses the $N_{ch}$-32-48-2 topology, and the 7 bins are temporally flattened as input, presented to the network as separate input timesteps. Each prediction still uses the membrane potential of the output neurons after input timesteps, and the network is reset for each prediction. Layer normalization is also applied on the SNN\_Flat inputs.

Figure \ref{fig:motor_result} shows plots of complexity and predictive quality of all four baseline networks. Both flattened networks demonstrate significantly greater $R^2$ performance than the other two networks. However, the larger input dimension of the ANN\_Flat network is reflected in its greater footprint, and the increased model timesteps and layer normalization sharply increase the effective operations of SNN\_Flat by two orders of magnitude compared to the simpler SNN. Thus, while input flattening and normalization increase the quality of model predictions for ANNs and SNNs, each comes with a significant complexity trade-off. 

\subsubsection*{Chaotic Function Prediction}

\begin{table}[ht]
\centerline{
\renewcommand{\arraystretch}{1.3}
\begin{tabular}{c|cccccccc} \thickhline
\multirow{2}{*}{Baseline} &
  \multirow{2}{*}{\begin{tabular}[c]{@{}c@{}}sMAPE\end{tabular}} &
  \multirow{2}{*}{\begin{tabular}[c]{@{}c@{}}Footprint\\ (bytes)\end{tabular}} &
  \multirow{2}{*}{\begin{tabular}[c]{@{}c@{}}Model Exec.\\ Rate (Hz)\end{tabular}} &
  \multirow{2}{*}{\begin{tabular}[c]{@{}c@{}}Connection\\ Sparsity\end{tabular}} &
  \multirow{2}{*}{\begin{tabular}[c]{@{}c@{}}Activation\\ Sparsity\end{tabular}} &
  \multicolumn{3}{c}{SynOps (per model exec.)} \\ \cline{7-9} 
        &       &          &    &     &       & \multicolumn{1}{c}{Dense}    & \multicolumn{1}{c}{Eff\_MACs} & Eff\_ACs \\ \hline
ESN & 14.79 & $2.81 \times 10^5$ & - & 0.876 & 0.0 & \multicolumn{1}{c}{$3.52 \times 10^{4}$} & \multicolumn{1}{c}{$4.37 \times 10^{3}$}  & 0 \\ 
LSTM  & 13.37 & $4.90 \times 10^5$ & - & 0.0 & 0.530 & \multicolumn{1}{c}{$6.03 \times 10^{4}$} & \multicolumn{1}{c}{$6.03\times 10^{4}$}  & 0 \\ \thickhline
\end{tabular}
}
\caption{Baseline results for the chaotic function prediction task. Execution rate is not reported as the data is a synthetic time series, with no real-time correlation.}
    \label{tab:MackeyGlass_resuls}
\end{table}

The chaotic function prediction task has two recurrent ANN baselines, which feature distinct network architectures:

\begin{itemize}
        \item \textbf{Long short-term memory (LSTM)} -- LSTMs are a class of recurrent ANN architectures~\cite{Hochreiter1997}, utilizing multiple gates for selective retention or omission of past information. The LSTM baseline consists of a single LSTM with a hidden state of 100 neurons, followed by a feed-forward layer to produce single-dimension output predictions. 
        In addition, the LSTM baseline utilizes explicit memory by buffering 50 previous datapoints, spatially flattening them into 50 input channels.
        
        \item \textbf{Echo state network (ESN)} -- ESNs are randomized recurrent ANNs that belong to a class of algorithms known collectively as reservoir computing~\cite{ScardapaneRandomness2017}, featuring more biologically-inspired principles than LSTMs despite not being spiking networks. Standard ESNs have only one hidden layer (the reservoir), where synaptic connections projecting input data to the hidden layer and recurrent synaptic connections within the hidden layer are chosen randomly and stay fixed during the training. 
        The model architecture for the ESN baseline has two neurons in the input layer, which projects the Mackey-Glass function input and additional constant bias input into a hidden layer of 186 neurons. Within the hidden layer, the probability of recurrent connections is set to 0.11. 

    \end{itemize}
The LSTM and ESN models were evaluated on a Mackey-Glass time series with $\tau=17$. The model is evaluated over 30 instantiations of the system; in each instance the start point is shifted forward by half of the Lyapunov time. The model is re-initialized and re-trained on each instance, and the results are averaged over all 30 instances. 


The ESN model is architecturally unique compared to the other ANN and SNN baselines. The connection sparsity metric ($0.876$) reflects the high number of zero-weight connections across its reservoir hidden layer. Due to this sparsity, hardware with support for sparse synaptic representation by ignoring zero weights would require less memory to represent the network, thus decreasing the deployed footprint of the model. The high connection sparsity of the ESN leads to significant reduction in synaptic operations - the ESN uses an order of magnitude fewer effective operations ($4.37\times10^3)$ than the LSTM ($6.03\times10^4$), while achieving comparable sMAPE. The activation sparsity of the ESN is 0 due to neurons using $\tanh(\cdot)$, rather than ReLU activations.

\begin{figure}[ht]
            \centering
            \includegraphics[width=.70\textwidth]{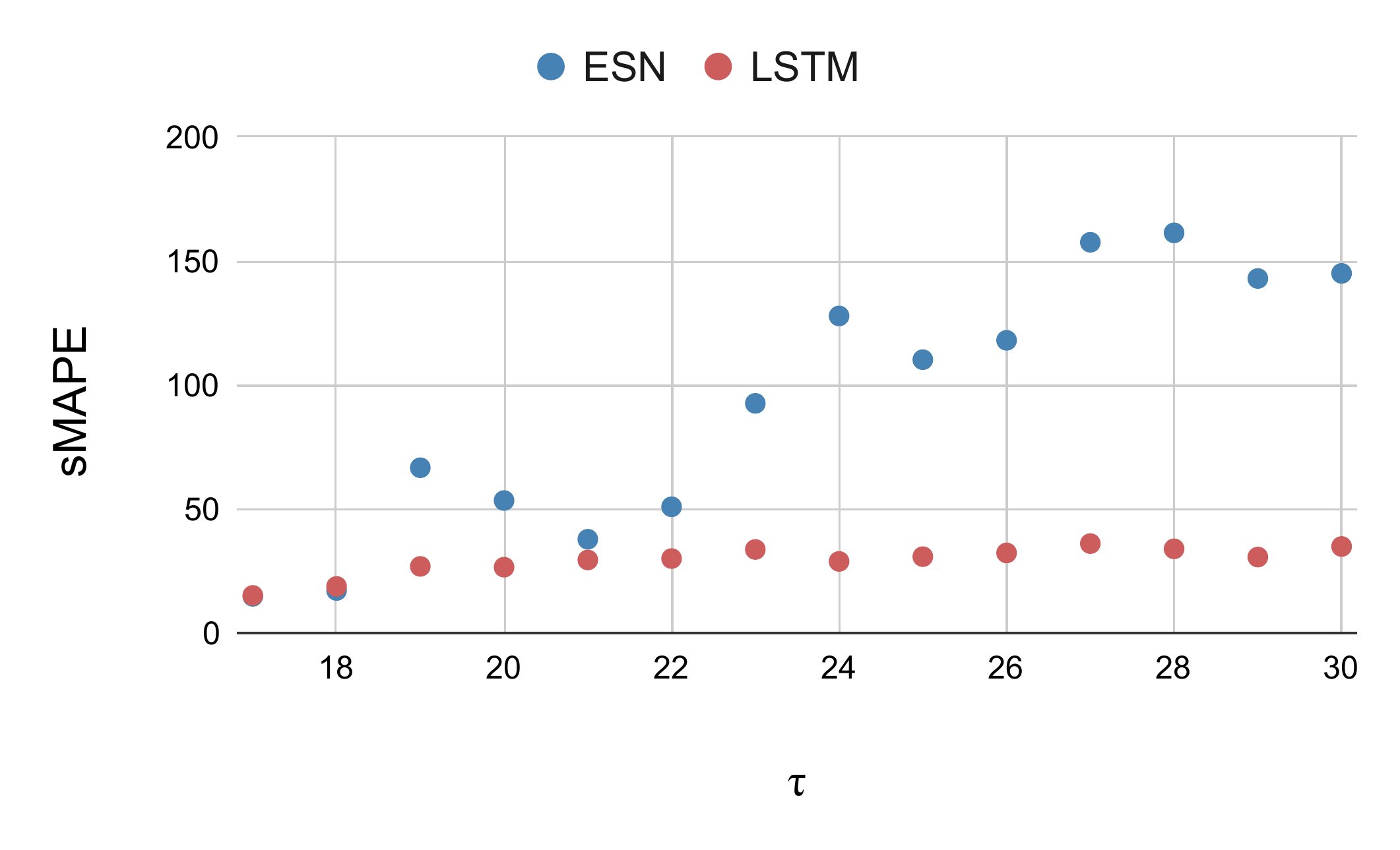}
            \caption{ESN and LSTM models evaluated on varying Mackey-Glass time series using a constant set of hyperparameters.}
        \label{fig:mackey_glass_function_performance}
    \end{figure}

Furthermore, we show the generalization and robustness capabilities of the particular ESN and LSTM models by applying them, with fixed hyperparameter sets, to other Mackey-Glass time series. Figure \ref{fig:mackey_glass_function_performance} shows the sMAPE score of the models over varied time series with the $\tau$ Mackey-Glass parameter varying between 17 and 30. The models were trained independently for each time series.
As the Mackey-Glass $\tau$ parameter characterizes the time-delay of the system, its increase roughly corresponds to prediction difficulty, shown by the increasing sMAPE trend through the plot. Notably, the LSTM maintains an error that is relatively lower than that of the ESN for all $\tau > 18$. However, the LSTM uses explicit memory via input buffering, so it is conjectured that the historical data allows for greater robustness to the varying time series characteristics. The ESN uses only one previous timestep, so its memory is only implicitly retained within its hidden layer. While the ESN tunes well to the $\tau$=17 case and demonstrates greatly reduced effective operations compared to the LSTM, the same set of hyperparameters does not generalize as well to other time series. Further research is motivated in explicit memory buffers versus implicit memory within the network state for trade-offs in single-series forecasting performance, complexity, and generalization capability.

\subsubsection*{Discussion and Opportunities for Further Research}
Baseline results for the four v1.0 algorithm track tasks compare the correctness and complexities of various solution types. Compared to ANNs, SNNs and ESNs demonstrate complexity advantages such as smaller footprints, high sparsity, and accumulate rather than multiply-and-accumulate operations. Especially on the motor prediction and chaotic function prediction regression tasks, the SNN and ESN baselines already achieve competitive correctness at lower complexity than the ANN and LSTM counterparts. Further research opportunities in model architectures, data pre-processing and buffering, and training paradigms to achieve greater performance is enabled by the standard framework and tooling provided by NeuroBench.
\section*{System Track Benchmark Framework}

While the algorithm track aims to benchmark solutions in a system-independent manner via complexity analysis, the NeuroBench system track aims to evaluate deployed execution time, throughput, and efficiency of systems comprised of an algorithm deployed and tailored to a hardware platform. \secondrev{Previous benchmark studies have examined neuromorphic systems under various applications, including keyword spotting~\cite{blouw2019, bos2024micropowerspokenkeywordspotting}, audio and video processing~\cite{shrestha24efficient}, and combinatorial optimization~\cite{chen2024onoffneuromorphicisingmachines, pierro2024solvingquboloihi2}. While these studies have demonstrated neuromorphic system advantages, the benchmark tasks have been unaligned. In order for the hallmarks of neuromorphic hardware to be aptly judged against conventional systems and foster the expansion of neuromorphic solutions, transparent and objective comparisons must be made on standard tasks between sufficiently mature neuromorphic systems head-to-head, as well as against conventional systems.}

    \begin{figure}[ht]
            \centering
            \includegraphics[width=.85\textwidth]{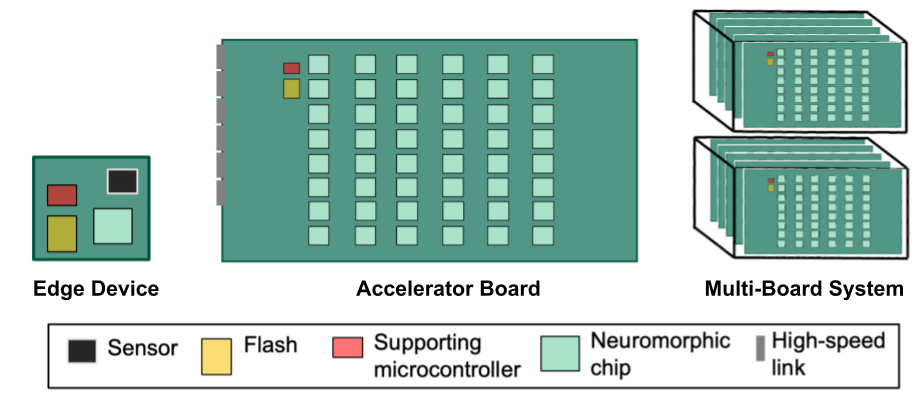}
            \caption{Types of neuromorphic systems at various integration scales.}
        \label{fig:system_types}
    \end{figure}

A key challenge for benchmarking neuromorphic hardware is that systems are implemented and deployed at vastly different scales to serve diverse applications, from cloud services (e.g., multi-chip platforms like Loihi~\cite{Davies18loihi} and SpiNNaker~\cite{mayr2019spinnaker}) to embedded sensing intelligence (e.g., Speck~\cite{synsensespeck} and SNP~\cite{InnateraSNP, Levy2021}). This range is visualized in Figure~\ref{fig:system_types}. \secondrev{Existing benchmarks for conventional systems have individual focuses across high-performance~\cite{top500}, datacenter-level computing~\cite{Reddi2020}, and embedded processing~\cite{banbury2021mlperf}. 
Thus, rather than pursuing a one-size-fits-all suite of tasks, the goal of the NeuroBench system track is to develop benchmarks at various scales and use cases, under multiple application areas in which both conventional and neuromorphic platforms may compete. The selected v1.0 NeuroBench system track benchmarks represent key commercial application areas for existing systems, and they differ from the tasks in the algorithm track, which are more research-oriented. As benchmark results continue to identify properties of highly effective algorithms and systems, the two tracks will converge to the same selction of tasks that are seen as the most impactful for future progress in the field.}

\secondrev{In this section, we present the system track guidelines outlining metrics and tasks, representing collective design between multiple owners and vendors of neuromorphic hardware. Baseline benchmark results for neuromorphic and conventional systems are reported, and further official results will be collected and announced at a regular cadence, akin to the MLPerf suite~\cite{mlperf_policies}. As with all other facets of the NeuroBench framework, the system track guidelines will continue to be adapted and extended iteratively as benchmark results are produced and shared. Up-to-date information on the latest benchmarks and official results can be found on the NeuroBench website (\url{https://neurobench.ai/}).}


    \subsection*{System Track Metrics}
    \secondrev{In order to be representative of the properties of a deployed system, the system benchmarks, like the algorithm benchmarks, are assessed at the task level for the overall system, as opposed to operation- or kernel-level assessment of individual components. Task-level benchmarks enable straightforward comparison between systems of any type with regard to their abilities to solve problems, and the overall system-level measurement describes the realistic capability and efficiency of a whole solution.}

    \secondrev{Each individual system benchmark uses task-specific metrics aligned with \textit{correctness}, \textit{timing}, and \textit{efficiency} to measure the system under test (SUT). The following general considerations are applied to each category:}
    \begin{itemize}
        \item \textbf{Correctness} -- \secondrev{In other system benchmarks, such as the closed category of the MLPerf Inference framework~\cite{Reddi2020}, the same trained model is used to benchmark all SUTs, and a correctness threshold is imposed to ensure optimizations such as lower precision do not disrupt task performance.}
        
        Due to the tight coupling between an algorithm and its system implementation in many existing neuromorphic hardware solutions, the particular model used to solve a NeuroBench system track benchmark task is unconstrained. Therefore, correctness must be measured to verify the validity of the solution. No correctness thresholds are imposed on submissions, but the benchmark leaderboard will impose tiers of solution correctness on submissions to evaluate accuracy-efficiency trade-offs of system approaches.
        
        \item \textbf{Timing} -- \secondrev{Depending on the task, timing performance can include measurements of sample throughput or execution time. Individually, the former entails an offline, batched inference benchmark, while the latter aligns with a streaming benchmark, in which one inference does not start until the previous one ends. Together, both throughput and execution time should be reported for tasks in which the SUT runs multiple inferences at any given time, each representing a request which must be responded to within a constrained window. The MLPerf Inference framework has defined widely-adopted general task scenarios corresponding to each of these categories (offline, single-stream, and server, respectively), and the NeuroBench system track will use these scenario guidelines where applicable to maintain consistency and build on conventional frameworks.}

        \secondrev{In addition, neuromorphic systems are also applied to tasks in which there is no notion of discrete sample throughput or execution, such as for heuristic approximations of intractable problems or operation over a continuous stream of data (e.g., from an event camera). Timing performance should be defined on a per-benchmark basis for such tasks, such as a time-to-solution latency or percentage of execution which exceeds a real-time threshold.}
        
        
        \item \textbf{Efficiency} -- 
        Conventional system benchmarks such as TOP500~\cite{top500} for HPC and MLPerf Inference~\cite{Reddi2020} for deep learning do not require power measurement submission in the main benchmark, instead allowing for separate submissions to an adjacent power track (Green500~\cite{green500} and MLPerf Power~\cite{mlperfpower}, respectively). Not only has efficiency been usually considered as a second-order metric for conventional systems, it is also notoriously difficult to precisely measure. However, as energy efficiency is a key hallmark of biology and thus is a focus of neuromorphic research, power and energy consumption must be first-order metrics in the NeuroBench system track. \secondrev{Similarly to timing metrics, efficiency metrics should be tailored on a per-benchmark basis, i.e., a real-time always-on processing task may focus on average power, while offline batched systems focusing on high-throughput inference may focus on both peak power and energy per inference.}
        
        
        
    \end{itemize}

    \secondrev{As neuromorphic systems currently explore a broad range of varied implementation approaches, board-level integration, and developmental stages of hardware, platform diversity poses difficult challenges for completely consistent hardware measurement methodologies (e.g., consistent power meters, chip interfaces, and data loading). Thus, to enable an initial step towards consistency in the system track while ensuring openness, we focus on the development of guidelines for transparent documentation, as they provide the foundation for shared methodology among highly diverse solutions. While there may be differences in how metrics are measured, salient details will be available to contextualize the results, allowing for holistic analysis, and leading the way for future consistency by enforcing transparency.}

    Benchmark submissions may perform separate runs to report performance and power in order to demonstrate system flexibility (e.g., a `performance-mode' run optimal for execution time and an `efficiency-mode' run optimal for energy), however in all runs, both metrics must be reported.

    Importantly for the NeuroBench system track, in measuring timing and efficiency, data pre- and post-processing must be taken into account. Neuromorphic methods will often consume and produce non-standard (e.g., event-based) data modalities, the processing of which may consume a significant amount of the overall execution time and may not be computed on the neuromorphic hardware itself. As many instances of neuromorphic hardware cannot be deployed without such associated processing, it is essential that measurements capture the cost of data processing, which stands in contrast with conventional system benchmarks whose measurements start from pre-processed data~\cite{mlperf_policies}.

    \subsection*{System Track Benchmarks}
    \secondrev{Two benchmark specifications for the v1.0 system track are defined in this article, covering embedded to datacenter scales. Full benchmark details are available in the Methods section.}
    \begin{itemize}
        \item \textbf{Acoustic Scene Classification} -- The acoustic scene classification benchmark challenges systems to classify audio into predefined categories based on the environmental audio context. 
        Such capabilities are key for hearable devices, which can utilize them to automatically adjust sound equalisation profiles, appropriately target microphone denoising, and support active noise cancellation.
        The application further challenges systems to fulfill technical requirements, such as always-on and real-time operation, and time series processing.
        Acoustic scenes provide a rich repertoire of features that are necessary for prediction, thus this task is a complement to keyword classification, which mainly focuses on shorter-term features (e.g., phonemes) with a relatively smaller feature repertoire.

        
        The benchmark evaluates the classification capabilities of both neuromorphic systems and conventional computing platforms using \revision{datasets from the DCASE challenge~\cite{Heittola2020}.} 
        These datasets consist of a myriad of audio recordings from diverse environments, including airports, public parks, and buses, thus providing a comprehensive foundation for testing both application- and system-level performance. \secondrev{The NeuroBench subset of the DCASE dataset includes 41360/16240 train/test samples across four classes (airport, street traffic, bus, park).}

        \secondrev{The task will be presented under the single-stream task scenario, providing one 1-second sample to the SUT at a time. Classification probability will be sampled to determine the correctness of the prediction. As the NeuroBench system track allows for unconstrained algorithmic implementation, pre-processing and inference metrics should be separately measured and reported together, which differs from prior system benchmarking that only measure inference~\cite{blouw2019, banbury2021mlperf}. 
        Timing results report on-device average execution time per sample.
        Since the platform diversity of edge-targeted systems poses inherent inconsistencies in efficiency measurement, power should be reported under idle and active contexts, following prior benchmark study methodology~\cite{blouw2019}. Idle power measures the system prepared for inference with the model loaded, and active power measures the system running pre-processing or inference. The difference between active and idle measurements offers dynamic power, which is used along with execution time to calculate dynamic energy-per-sample.}
        

        \item \textbf{QUBO} -- 
        As a non-ML task, NeuroBench incorporates quadratic unconstrained binary optimization (QUBO). QUBO is a particularly beneficial first optimization task for NeuroBench for two reasons. First, the binary variables are a natural fit for neuromorphic systems with purely binary spike communication. Second, real-world QUBO applications typically feature sparse cost matrices~\cite{koch2022} which benefit from the sparse synaptic connectivity and execution that neuromorphic systems are often optimized for \cite{Aimone2022}. The initial set of QUBO workloads in NeuroBench searches for the maximum independent (i.e. unconnected) set of nodes in graphs, a task that has wide applications across industry and academia, such as resource allocation in wireless networks, portfolio optimization, and task scheduling~\cite{wurtz2024industryapplicationsneutralatomquantum}.
        
        NeuroBench provides a QUBO generator that can uniquely specify each workload by three specific parameters provided by the benchmark: the number of graph nodes, the density of graph connections, and a random seed. The generator provides a large dataset for reliable statistics and allows scaling from modest workloads for small-scale and prototype systems to large workloads for larger-scale systems. \secondrev{The graph sizes specified by the benchmark increase in a pseudo-geometric progression (10, 25, 50, 100, 250, ...), and submissions are encouraged to extend the problem size to the limits of the SUT. Graph density ranges from 1\% to 30\%, in order to show the relationship between the number of connections and SUT power. 5 random seeds for each setting should be tested.}


        \secondrev{Optimization algorithms use heuristic methods to iteratively refine approximate solutions to intractable problems that cannot be completely solved. The QUBO benchmark thus measures solution optimality and energy consumption after fixed, pre-set runtimes, removing any timing measurement. Solution optimality is defined as BKS-Gap, a relative gap between the current SUT's solution compared against the best-known solution (BKS) to the same problem found using a high-powered solver with a long runtime.
        }
        

    
    \end{itemize}

\subsection*{Baseline Results}

\secondrev{Baseline results for each of the two system track benchmarks are provided for a mature neuromorphic system against a conventional platform. Like for the algorithm track baseline results, the system track baselines are intended to snapshot the solution space and provide starting points for the task leaderboards. Further details on each baseline system are available in the Methods section, and in-depth system documentation for the Xylo ASC baseline and CPU/Loihi 2 QUBO baselines are provided by Ke et al.~\cite{ke2024neurobenchdcase2020acoustic} and Pierro et al.~\cite{pierro2024solvingquboloihi2}, respectively.}

\subsubsection*{Acoustic Scene Classification}

\secondrev{For the acoustic scene classification task, two baseline embedded systems are reported:
\begin{itemize}
    \item \textbf{CPU} -- The CPU baseline is an Arduino Nano 33 BLE, which uses an ARM Cortex M4 microcontroller for both pre-processing and inference. The digital audio sample is pre-processed using Mel-filterbank energies (MFE), and inference uses a conventional CNN. Execution time is measured using on-chip timers, and power is measured using total system power.
    \item \textbf{Xylo} -- The Synsense Xylo~\cite{synsensexylo} neuromorphic baseline uses a feed-forward SNN with multiple synaptic time constants~\cite{bos2024micropowerspokenkeywordspotting}. As the system is intended for continuous, real-time audio processing, the board uses an analog front end to pre-process analog audio signals directly from a microphone into spikes for the digital inference engine. To conform with a digital benchmark dataset, a simulator of the analog pre-processor generates spikes, which are routed to the inference module. Execution time and power are measured using on-board instruments.
\end{itemize}}

\begin{table}[ht]
\centerline{
\renewcommand{\arraystretch}{1.3}
\begin{tabular}{@{}cc|cccccc@{}}
\toprule
\multicolumn{2}{c|}{\multirow{2}{*}{Baseline}}                        & \multirow{2}{*}{Accuracy} & \multirow{2}{*}{\begin{tabular}[c]{@{}c@{}}Execution\\ Time (ms)\end{tabular}} & \multirow{2}{*}{\begin{tabular}[c]{@{}c@{}}Idle \\ Power (mW)\end{tabular}} & \multirow{2}{*}{\begin{tabular}[c]{@{}c@{}}Active\\ Power (mW)\end{tabular}} & \multirow{2}{*}{\begin{tabular}[c]{@{}c@{}}Dynamic \\ Power (mW)\end{tabular}} & \multirow{2}{*}{\begin{tabular}[c]{@{}c@{}}Dynamic\\ Energy (mJ/inf)\end{tabular}} \\
\multicolumn{2}{c|}{}                                                 &                           &                                                                         &                                                                             &                                                                              &                                                                                &                                                                                    \\ \midrule
\multicolumn{1}{c|}{CPU}                       & \textit{pre-process} & \multirow{2}{*}{79.64\%}  & 43                                                                      & 79.40                                                                       & 100.72                                                                       & 21.32                                                                          & 0.917                                                                              \\
\multicolumn{1}{c|}{\textit{(system-wise)}}    & \textit{inference}   &                           & 45                                                                      & 79.40                                                                       & 100.15                                                                       & 20.75                                                                          & 0.934                                                                              \\ \midrule
\multicolumn{1}{c|}{Xylo}                      & \textit{pre-process*} & \multirow{2}{*}{79.90\%}  & -                                                                       & 0.00017*                                                             & 0.015*                                                                        & 0.015*                                                               & 0.015*                                                                                  \\
\multicolumn{1}{c|}{\textit{(component-wise)}} & \textit{inference}   &                           & 84                                                                      & 0.351                                                                       & 0.692                                                                        & 0.341                                                                          & 0.028                                                                             \\ \bottomrule
\end{tabular}
}
\caption{\secondrev{Baseline results for the acoustic scene classification task. Pre-processing of the Xylo (marked with an asterisk *) measures power of the analog pre-processor in real-time relative to the audio data, whereas other measurements are of digital components processing digital data on-hand. Idle Xylo pre-processing measures silence and active measures test data audio played to the device, and energy is measured over the sample duration of 1 second. CPU power is measured over the full Arduino system, as it does not have on-board power instrumentation, while Xylo power measures power consumed by the Xylo Audio 2 ASIC only. Dynamic power and energy provide proper comparison between the systems, and idle and active measurements are provided for transparency.}}
\label{tab:asc_results}
\end{table}

\secondrev{Table~\ref{tab:asc_results} lists baseline results for the neuromorphic Xylo system against an Arduino system. Compared to prior neuromorphic audio system benchmarking~\cite{blouw2019}, which takes a server-class CPU as a point of comparison, we adopt a fairer approach by focusing on low-power edge application and comparing against an Arduino embedded microprocessor. At comparable inference accuracy, Xylo exhibits $60.9\times$ less dynamic inference power and $33.4\times$ less dynamic inference energy consumption than the Arduino.
}


\subsubsection*{QUBO}

\begin{figure}[t]
    \centering
    \begin{subfigure}{0.48\textwidth}
        \includegraphics[width=\linewidth]{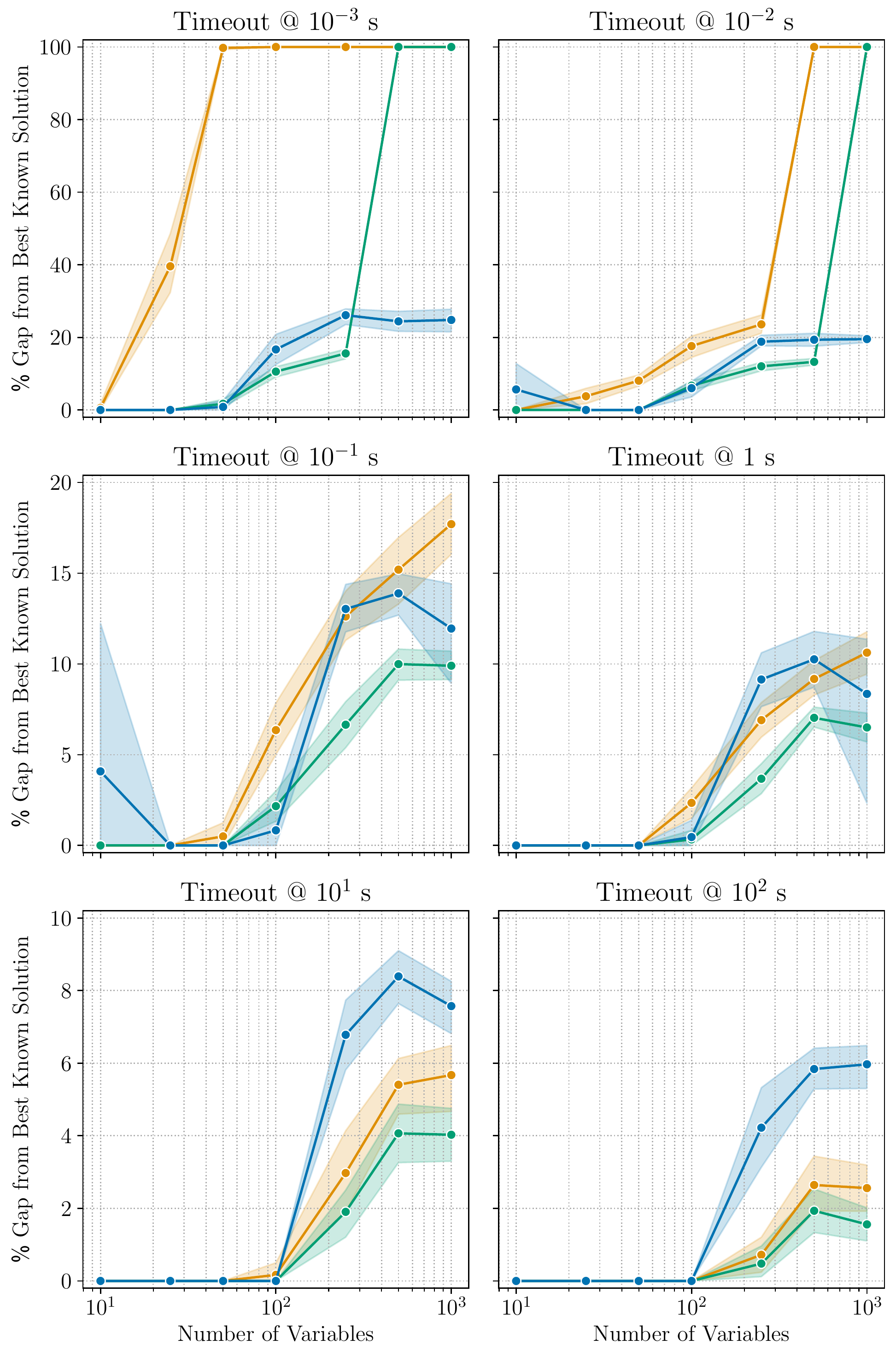}
    \vspace*{0.2cm}
    \end{subfigure}\\
    \begin{subfigure}{0.3\textwidth}
        \includegraphics[width=\linewidth]{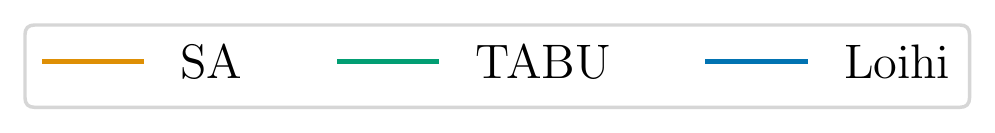}
    \end{subfigure}
    \caption{\secondrev{Percentage gap from the best known solution (BKS-Gap\%) for the QUBO workloads with QUBO matrices at $15\%$ density (lower is better). Results are shown for different timeouts of the QUBO solvers. Figure taken, with permission, from Pierro et al.~\cite{pierro2024solvingquboloihi2}.}}
    \label{fig:qubo_bksgap}
\end{figure}

\begin{figure}[t]
    \centering
        \includegraphics[width=\linewidth]{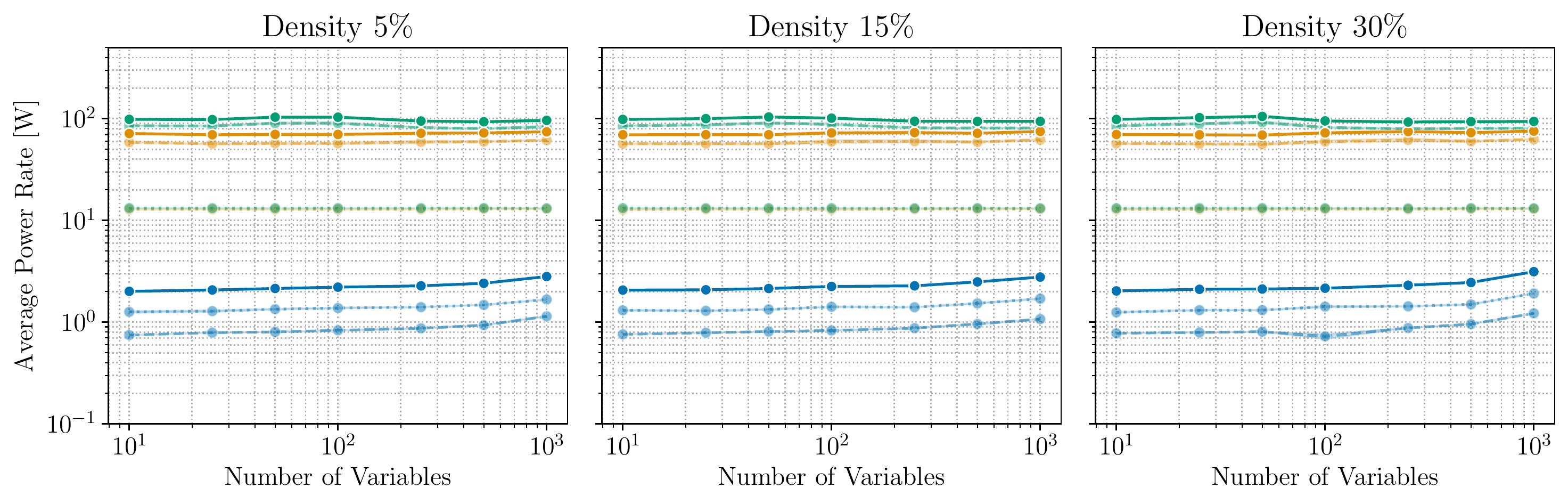}
    \vspace*{0.5cm}
\includegraphics[width=0.3\linewidth]{figures/legend.pdf}
\includegraphics[width=0.35\linewidth]{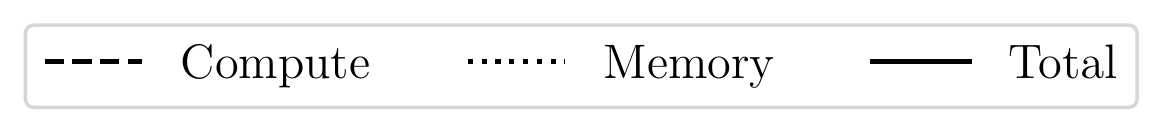}
        \caption{\secondrev{Power consumption of the QUBO solvers running simulated annealing (SA) or TABU search on CPU, and the parallelized version of simulated annealing on Loihi 2. The CPU solvers require up to $37\times$ more power than the neuromorphic algorithm on Loihi 2. Since the QUBO workloads were run for a fixed timeout, differences in power consumption are equivalent to differences in energy consumption of the processors. Figure taken, with permission, from Pierro et al.~\cite{pierro2024solvingquboloihi2}.}}
        \label{fig:qubo_power}
\end{figure}

\secondrev{Three baselines are measured for the QUBO benchmark:
\begin{itemize}
    \item \textbf{Simulated Annealing (SA)} -- The simulated annealing solver uses Markov chain Monte Carlo (MCMC) sampling to probabilistically explore the search space. 
    \item \textbf{Tabu Search (TABU)} -- Tabu search solvers maintain and iterate a list of prohibited actions in order to prevent the search from remaining in local minima or revisiting states. Both the TABU and SA solver baselines use the D-Wave Samplers library~\cite{dwavesamplers} on an Intel Core i9-7920X desktop-class CPU, with power measured using Intel SoC Watch.
    \item \textbf{Loihi 2} -- The Loihi 2 neuromorphic system solver uses an SNN formulation of the simulated annealing algorithm which enables solving via neural dynamics, and parallelization via stochastic refractory periods. The baseline is implemented on one Loihi 2 chip on the 8-chip Kapoho Point board, and internal power instrumentation measures all compute and memory components of the chip.
\end{itemize}}

\secondrev{Figure~\ref{fig:qubo_bksgap} shows the optimality reached by the CPU- and Loihi 2-based solvers after different timeouts. For tight time constraints, at $10^{-2}$ seconds timeout or less, Loihi 2 finds feasible solutions to workloads $4\times$ larger than the CPU. But for timeout lasting $10$ seconds or longer, the CPU running TABU provides the lowest BKS-Gap, incentivizing algorithmic advances for neuromorphic optimization systems.}
\secondrev{Figure~\ref{fig:qubo_power} illustrates the power consumed during runtime. Across the workloads, the Loihi 2 solver requires $37.24\times$ less power compared to the best CPU solver.  
}

\subsubsection*{Discussion and Future Work}
\secondrev{The initial baselines for the v1.0 system track compare correctness, timing, and efficiency of neuromorphic systems against conventional CPU systems in domains of both audio classification and optimization. Against mature, commercially-developed CPU systems, for both edge and server use cases, the neuromorphic systems show strong advantages in general efficiency, as well as further promises in terms of timing and correctness. 
}

\secondrev{In future NeuroBench iterations, the system track benchmarks can be unified under common tooling, similar to the algorithm track. Software toolchains such as Lava~\cite{lava}, Fugu~\cite{aimone2019composing}, SPyNNaker~\cite{spynnaker}, and Samna~\cite{samna}, among others, have been developed to interface with specific hardware platforms. Many of the stacks are built with general paradigms to support extension to any backend, and the community is actively moving towards developing standards for deployment tools. The current v1.0 benchmark specifications allow for open algorithm and software design in order to demonstrate fully optimized performance for neuromorphic systems. As standards mature in the future, a core focus of the NeuroBench system track is to introduce a closed-algorithm benchmarking category that leverages the recently proposed NIR model description framework~\cite{NIR2024} as a general, cross-platform tool for benchmarking key workloads of interest across many different platforms.}

\section*{Discussion} 

Benchmarking neuromorphic computing has faced challenges stemming from the diversity of neuromorphic approaches, the range of implementation and deployment tools, and rapid research evolution.
NeuroBench addresses these challenges as a framework for the inclusive, actionable, and iterative benchmarking of neuromorphic solutions, by including novel tasks and metrics, open-source and extendable harness tooling, and facilitating systematic growth via community collaboration. NeuroBench is supported and developed by a broad community of neuromorphic researchers to be a standard, agreed-upon benchmarking framework for neuromorphic technology.

Initial NeuroBench benchmarks span applications across domains of continual learning, computer vision, sensorimotor prediction, and time-series forecasting, as well as system implementations for audio and optimization settings. Baselines for each benchmark of the complete v1.0 algorithm track demonstrate the utility and validity of the metric framework, and offer a starting point to further algorithmic research in model architecture and training for greater performance and lowered complexity. 
\secondrev{The system track v1.0 benchmark baselines similarly provide the groundwork for both conventional and head-to-head comparisons of deployed hardware systems. Via collaboratively-defined measurement methodology and task specifications, the shared system track guidelines will be modified and extended to define standard protocols for continuous-time execution, analog circuit implementations, and other exploratory platforms in simulation stages, such as memristive hardware.}


Another important direction for NeuroBench is towards closed-loop benchmarks~\cite{closedloop-andre, closedloop-terry}. Biological systems excel in interacting with dynamic environments, demonstrating high energy efficiency, real-time reaction, and versatility. As such, embodied intelligence with adaptive sensory and action capabilities are of interest to neuromorphic research. In closed-loop scenarios, the objective is to sense and act within an environment to complete a task, rather than to statically process a frozen dataset, thus the benchmark harness infrastructure and measurement protocols will be extended to facilitate such benchmarks.



All future NeuroBench expansion will be informed by collected results and continue to be driven by the interests and development of the broader community.

\section*{Methods}
\label{sec:methods}


This section outlines details and specifications of the benchmark metrics, tasks, and baselines.

\subsection*{Algorithm Track Metrics}


NeuroBench includes correctness and complexity metrics, the latter of which is divided in static and workload metrics. Static metrics do not depend on the model inference and input data, while the workload metrics do. Note that the defined metrics reflect only the model and model execution. Data pre-processors and post-processors are not taken into account in the v1.0 algorithm track results. 

\subsubsection*{Footprint}
The footprint metric reflects the memory footprint a model. It is distinct from execution memory, which may incur further usage, e.g. to store activations. It is computed for a model by accumulating the sizes of the model's parameters and buffers, in bytes.
Parameters store the model synaptic weights, and buffers include other inference memory requirements, such as the internal states of recurrent or spiking layers and buffers of recent input data, if the model must record data for input binning. Considering $n$ parameters, each requiring $p_i$ bytes, and $b$ buffers of size $q_j$, the total model footprint is $\sum_{i=0}^{n}p_i + \sum_{j=0}^{b}q_j$.

\subsubsection*{Model Execution Rate}
Execution rate is a numeric which is not directly computed by the harness, but should be reported by the user. 
The numeric reflects the real-time correlation of the rate at which the model computes input data. If the model processes input with a temporal stride of $t$~seconds, then the rate should be reported as $t^{-1}$~Hz. Note the distinction between \textit{stride} and \textit{bin window} - input can be binned in overlapping windows, but execution rate depends on the temporal stride of window processing. As an example, a model may use 50~ms windows of input and compute every 10~ms, which would give an execution rate of 100~Hz.

This numeric is currently not well-defined for models operating under event-based or continuous-time contexts. These limitations will be addressed in future benchmark versions.

\subsubsection*{Connection sparsity} 
The parameter matrices of each layer $l$ in a model, representing synaptic weights, are collected, and the number of zero weights $m_l$ and total weights $n_l$ are aggregated, with the connection sparsity defined as $\frac{\sum_l m_l}{\sum_l n_l}$.



\subsubsection*{Activation sparsity}
Activation sparsity is computed after the inference phase. The sparsity is calculated by accumulating the number of zero activations ($z$), over all neuron layers ($l$), timesteps ($t$), and input samples ($i$) and dividing by the total number of neurons ($N$), $\frac{\sum_l \sum_t \sum_i z_{l,t}^i}{\sum_t \sum_l \sum_i N_{l,t}^i}$. 
The outputs of ReLU functions and spikes from spiking neurons are considered activations.

\subsubsection*{Synaptic operations}
Synaptic operations are the multiplication of weights by activation or input data, and are calculated using the inputs and weights of connection layers (e.g., \texttt{torch.nn.Linear} and \texttt{torch.nn.Conv2d}).
Effective synaptic operations are operations where a non-zero weight is multiplied by a non-zero activation. 
Effective operations are further divided into multiply-accumulates (MACs), and accumulates (ACs), where accumulates correlate with activations or input data only containing values of [-1, 0, 1], and multiply-accumulates cover all other cases.
The reported number of synaptic operations is the average number of synaptic operations required per model execution, the rate of which is defined by the \textit{model execution rate} metric.



The number of \textit{effective} synaptic operations is computed by performing the forward pass of a layer and counting the number of operations in which there is no zero multiplication.
Practically, this is implemented in the harness by setting all non-zero weights in the layer and all the non-zero activations to 1, then performing the forward pass and summing the output to give the number of synaptic operations. 


The number of \textit{dense} synaptic operations is computed in a similar fashion, by setting all weights and activations to 1 and accumulating the output of the forward pass.
Biases are not taken into account in the calculation of the synaptic operations, as they are added after weight multiplications and accumulation. 

Note that processing of activations before the connection layer, for instance using batch normalization, can transform sparse activations into dense input at the connection layer, which will lead to high effective synaptic operations despite high activation sparsity. Furthermore, such processing can transform binary activations to non-binary data, causing effective operations to be MACs rather than ACs. When deployed to neuromorphic hardware, such algorithms that normalize activations before multiplication with synaptic weights may lose the benefits of sparse operation, e.g., an SNN with normalization following each spiking layer would require dense MAC weight calculation, no matter how few spikes were generated.

In some cases, algorithm execution may have distinct temporal sections of higher and lower synaptic operations, such as during initial caching versus continuous inference. For such algorithms, benchmark users may choose to distinguish synaptic operations and other complexity measurements between execution sections.



\subsection*{Algorithm Track Benchmark Tasks}
\subsubsection*{Keyword FSCIL}
Few-shot Class-Incremental Learning, FSCIL, is an established benchmark task setting in the computer vision domain~\cite{tao2020fscil}. It can be defined as follows: a base session with fixed classes, each with abundant training data, is used to train an initial model. Then, successive incremental training sessions introduce new classes in a few-shot learning scenario. In each session, only the current session classes are available to the model for training. After each incremental training session, the model is evaluated on all previously seen classes, including the base classes. Therefore, the model has to learn new classes while retaining knowledge about the previously learned ones. 

Formally, for $M$-step FSCIL, where $M$ is the total number of incremental sessions, each training session uses a support dataset $D^{(t)}$, $t\in [0,M]$ to train new classes on. ${L^{(t)}}$ is the set of classes of the $t$-th session where $\forall i,j$  where $i \neq j$, $L^{(i)}\cap L^{(j)}=\varnothing$, meaning each training session uses a unique set of classes. $D^{(0)}$ and $L^{(0)}$ are the base class training data and set of base classes, respectively, $D^{(1)}$ and $L^{(1)}$ represent the first incremental session set, and so on. At session $t$, only $D^{(t)}$ is available for training, and for $t>0$, $D^{(t)}$ contains a fixed number of classes ($N$) with few samples per class ($K$). This form of FSCIL is therefore named $N$-way $K$-shot FSCIL. At the end of each session $t$, model accuracy is reported on the test samples of all previously seen classes $\{L^{(0)} \cup L^{(1)} \cup ... \cup L^{(t)}\}$.


For the Keyword FSCIL task, classes in the base set ($L^{(0)}$) have 700 samples each, with a fixed train/validation/test sample split of 500/100/100. All classes within incremental sessions have 200 samples per word, with a fixed train/test split of 100/100. Of the 100 training samples, 5 are randomly selected for few-shot learning (each session is 10-way, 5-shot). The inclusion of 200 samples allows for increasing learning up to 100 samples.

NeuroBench proposes an audio keyword classification version of the FSCIL task, which to the best of our knowledge is the first of its kind. This novel task is established by selecting a subset of the words and languages from the Multilingual Spoken Word Corpus (MSWC)~\cite{mazumder21mswc} dataset. The FSCIL task consists of a multilingual set of 100 base classes and 10 incremental sessions of 10 classes each, for a final total of 200 learned classes. 
Fifteen languages are represented: the base classes are composed of a set of five base languages with 20 words each, and each of the ten incremental sessions contains 10 words from a distinct language. The languages were chosen based on data availability within the MSWC dataset. The top five languages with the greatest number of potential words (words with enough data samples) are used as the base class languages, while the next ten languages with the greatest numbers are the incremental classes. 
The base languages are English, German, Catalan, French and Kinyarwada. Incremental languages are Persian, Spanish, Russian, Welsh, Italian, Basque, Polish, Esparanto, Portuguese and Dutch. The order of languages presented in the incremental sessions are randomized, but each incremental session will represent exactly one new language.

For each language, the longest length words (that had the appropriate number of samples) were selected to allow for rich and robust temporal features to be learned. Next to the richness of longer words, there are practical considerations for this choice. The MSWC dataset normalizes all samples to a duration of 1 second, centered around the 0.5 seconds mark. For shorter words, this means that the data needs to be zero-padded on both sides to fill the entire duration. Longest-length words are likely to fill the complete sample and reduce zero-padding, which is also useful in scenarios in which algorithms seek to classify words before the sample has completed~\cite{jeffares2022spikeinspired}. Furthermore, common keyword spotting solutions, such as \textit{Ok Google}, \textit{Alexa}, and \textit{Hey Siri}, use multi-syllable wake-phrases to assist in accurate word classification. \revision{Using shorter keyword subsets can lead to greater challenge in both base language training and continual class learning. The full list of chosen words for the task presented in this paper (longest words), as well as other potential subsets of short words for the same languages, can be found with dataset documentation through the harness.}

Within each language, words showing great similarity in phonics and meaning are not included (e.g. l'amendement and amendements in French). Across different, but related languages, words with similar pronunciation and meaning were not included as well (e.g., university, universität and universitat in English, German and Catalan).

The subset of MSWC used for this FSCIL task is significantly smaller in size (630MB) compared to the full MSWC datatset (124GB), and subset download details can be found in the harness.


\subsubsection*{Event Camera Object Detection}
The task of object detection using event camera data involves identifying bounding boxes of objects belonging to multiple predetermined classes in an event stream. The dataset is the Prophesee 1 Megapixel automotive detection dataset~\cite{Perot2020}, which is one of the largest and highest-resolution event-camera detection datasets currently available. The performance of the task is defined by the COCO mean average precision (mAP) metric~\cite{LinCOCO_2014}, a metric that is commonly used for the evaluation of object detection algorithms. Only three out of the seven available object classes within the dataset are used due to limited sample availability in the dataset, which matches prior work~\cite{Perot2020}.

COCO mAP is calculated using the intersection over union (IoU, Equation~\ref{eq:iou}) of the bounding boxes produced by the model against ground-truth boxes. Here, $A$ and $B$ refer to bounding boxes, and the intersection and union consider the overlapping area and the area covered by both boxes, respectively. The IoU is compared against 10 thresholds between 0.50 and 0.95, with a step size of 0.05.
For each threshold, precision is calculated (Equation~\ref{eq:precision}) with True Positives (\textit{TP}) and False Positives (\textit{FP}) determined by whether the IoU meets the threshold or not, respectively. The mAP is calculated as the averaged precision over all thresholds for each class, which is further averaged over all classes to produce the final result.

\begin{equation} \label{eq:iou}
    IoU(A,B) = \frac{|A\cap B|}{|A\cup B|}
\end{equation}

\begin{equation} \label{eq:precision}
    Precision(TP,FP) = \frac{\sum TP}{\sum TP + \sum FP}
\end{equation}



Note that in the dataset, labels are generated from images from an RGB camera. Due to the nature of event cameras, objects which are still at the start of a recording sequence have no generated events and cannot be detected. Therefore, labels within the first 0.5 seconds of each sequence are not taken into account. Furthermore, as the RGB camera used for labeling has a higher resolution than the event camera, not all objects which appear in the RGB image are recognizable from the generated events. Thus, objects with a diagonal of less than 60 pixels are also not considered. The dataset and metric measurement is implemented using the Prophesee Metavision software~\cite{metavision}.


\subsubsection*{Non-human Primate Motor Prediction}
The non-human primate motor prediction task involves predictive modeling of two-dimensional fingertip velocity, given neural motor cortex data.
The six sessions used for the benchmark comprise three recording sessions each from two non-human primates (NHP Indy and NHP Loco) such that the chosen sessions approximately span the entire duration of the experiment~\cite{makin-dataset} (several months).
The specific sessions used are indy\_20170131\_02, indy\_20160630\_01, indy\_20160622\_01, loco\_20170301\_05, loco\_20170215\_02, and loco\_20170210\_03.
Each of these sessions consists of one day of experiments, during which multiple reaches are recorded. During each reach, a target position is displayed, which the NHP needs to localize and touch with its finger. Once the NHP touches the correct target for the current reach, the next reach is instantiated, showing a new target position. The data contains sensorimotor cortex recordings from 96 channels for the recordings of the first NHP (Indy), while 192 channels were used for the second NHP (Loco), and was gathered and labeled at a frequency of 250Hz. Two-dimensional position data of the NHP fingertip during its reaches is provided in the dataset, and these are translated into $X$ and $Y$ velocity ground-truth labels using discrete derivatives~\cite{makin-paper}.

Each session is segmented into individual reaches based on the target position for the NHP to touch. 
The data in each session is split such that the initial $75\%$ of reaches are used for training and validation, and the remaining $25\%$ of reaches are test data. The user can choose how to utilize the training and validation split for their particular method.  

During evaluation, the coefficient of determination ($R^2$, Equation~\ref{eq:r2}) for the $X$ and $Y$ velocities are averaged to report the correctness score for each session, where $n$ is the number of labeled points in the test split of the session, $y_i$ is the ground-truth velocity, $\hat{y_i}$ is the predicted velocity, and $\bar{y}$ is the mean of the ground-truth velocities. The $R^2$ from sessions for each NHP are averaged, producing two final correctness scores.

\begin{equation} \label{eq:r2}
    R^2 = 1 - \frac{\sum_{i=1}^{n} (y_i - \hat{y_i})^2}{\sum_{i=1}^{n} (y_i - \bar{y})^2}
\end{equation}

\subsubsection*{Chaotic Function Prediction}

The chaotic function prediction is another sequence-to-sequence problem. Given an input sequence generated from a one-dimensional Mackey-Glass function, the task is to predict the future values of the same function.
The dataset used for this task is synthetically generated, following the Mackey-Glass differential equation~\cite{mackey1977oscillation} (Equation \ref{eq:MackeyGlass}), which is integrated and discretized with a timestep of $\Delta t$. The time series generated by this differential equation is a function of the Mackey-Glass parameters $n$, $\beta$, $\gamma$ and $\tau$. Adhering to standard parameters~\cite{jaeger2004harnessing}, the values used for $n$, $\beta$, and $\gamma$ are 10, 0.2 and 0.1 respectively. $\tau$ is varied between 17 (a standard value) and 30, leading to 14 time series which vary greatly in dynamics and can be used to analyze the generalization of predictive models.

Each value of $\tau$ is associated with a Lyapunov time, the expected predictability timescale for chaos~\cite{gilpin2023largescale}, which is used as the time unit for each series. To calculate the overall Lyapunov time for each value of $\tau$, we average the Lyapunov times of 30,000 generated time series of 2,000 timesteps, with $\Delta t=1.0$, each with a randomly chosen initial condition.
All time series and Lyapunov times were generated and estimated using the \texttt{JiTCDDE} library~\cite{jitcxde}. 
For each final time series used for benchmarking, initial conditions are a point randomly chosen along the series.
The Lyapunov time and initial condition $x_0$ for each of the 14 final time series are provided in Table \ref{tab:MG_params}.


\begin{equation} \label{eq:MackeyGlass}
    \frac{dx}{dt} = \frac{\beta x(t-\tau)}{1 + x(t-\tau)^n} - \gamma x(t)
\end{equation}

As the integration of the differential equation can depend on underlying floating-point arithmetic and thus produce varying time series on different machines, the datasets are precomputed and loaded for training and evaluation. In the benchmark results, 30 instantiations of the Mackey Glass system are used, each with a length of 20 Lyapunov times and successively shifted forwards by half a Lyapunov time. The dataset time series are generated for 50 total Lyapunov times to allow for varied offset starting points. The generated time series are available to be downloaded under the NeuroBench harness.

\begin{table}[ht]    
    \begin{center}
    \begin{tabular}{|c|c|c|}
        \hline
        $\tau$ & Lyapunov Time & $x_0$ \\ \hline
        17 & 197 & 0.7206597 \\
        18 & 138 & 0.7744313 \\
        19 & 315 & 0.7783468 \\
        20 & 131 & 0.9225991 \\
        21 & 191 & 0.9479431 \\
        22 & 119 & 0.5455960 \\
        23 & 106 & 0.8622247 \\
        24 & 97 & 0.3259660 \\
        25 & 98 & 0.8297825 \\
        26 & 104 & 1.0033490 \\
        27 & 112 & 0.6491406 \\
        28 & 119 & 1.0957495 \\
        29 & 131 & 0.9256179 \\
        30 & 139 & 0.2713639 \\
        \hline
    \end{tabular}
    \end{center}
    \caption{Mackey-Glass parameters used for the 14 time series.}
\label{tab:MG_params}
\end{table}

Symmetric mean absolute percentage error (sMAPE, Equation~\ref{eq:smape}), a standard metric in forecasting~\cite{MAKRIDAKIS202054}, is used to measure the correctness of the model predictions $\hat{y_i}$ against the ground-truth $y_i$, over $n$ data points in the test split of the time series. 
The sMAPE metric has a bounded range of $[0, 200]$, thus diverging predictions (infinity or NaN) due to floating-point arithmetic have bounded error which can be used to average correctness over multiple time series instantiations.

\begin{equation} \label{eq:smape}
    sMAPE = 200 \times \frac{1}{n} \left( \sum_{i=1}^{n} \frac{|y_i - \hat{y_i}|}{(|y_i| + |\hat{y_i}|)}\right)
\end{equation}

\subsection*{Algorithm Track Baselines}

All baselines are implemented using PyTorch \texttt{nn.Module} objects in order to interface with the harness.

\subsubsection*{Keyword FSCIL}

The ANN baseline employs Mel-frequency cepstral coefficients (MFCC) pre-processing along with a modified version of the M5 deep convolutional network architecture~\cite{dai2016deep}. 

The MFCC pre-processing converts the 48~kHz, 1~second audio samples from MSWC into 20 channels of 200 timesteps (5~ms stride, 10~ms time bins), focusing on frequencies within the human voice range between 20~Hz and 40~kHz.
The network contains four successive blocks, each consisting of 1D convolution, batch-normalization, ReLU activation, and max-pooling layers, followed by a single readout fully-connected layer. Convolutional layers apply their kernels over the temporal dimension of the samples, thus extracting longer temporal features through the depth of the network. 
We also incorporate dropout after the ReLU activations to avoid over-fitting and let the network be more general for incremental learning. 
The network is trained with stochastic gradient descent using cross-entropy loss and the Adam optimizer.

For the SNN baseline, we employ the Speech2Spikes~\cite{stewart23speech2spikes} (S2S) preprocessing algorithm to convert audio samples to spikes.
For S2S we use the default parameters from the original implementation, only the hop length is updated to match the 48~kHz audio frequency of the MSWC samples, whereas the original implementation was applied to 16~kHz audio. S2S applies a Mel Spectrogram and a log operation to raw audio samples, converting them to positive and negative trains of spikes using delta-encoding.

Spike trains from S2S are used as input for the recurrent SNN (RSNN), which consists of 2 recurrent adaptive leaky integrate-and-fire (RadLIF) layers of 1024 neurons and one linear output layer. The model architecture is adapted from Bittar's work~\cite{bittar22surrogate}.
The RadLIF neurons in these layers are LIF neurons that produce a binary spike $\mathbf{s}(t)$ and reset via subtraction when their membrane potential $\mathbf{u}(t)$ crosses a certain threshold value $\theta$, combined with an extra adaptation variable $\mathbf{w}(t)$ to enable more complex temporal dynamics and firing patterns. Equation~\ref{eq:adLIF_current} is the input current to neurons, with $x$ the input spikes from the previous layer, $W_f$ the forward weight matrix, $BNTT$ batch-normalization through time, and $W_r$ the recurrent weight matrix. $\mathbf{u}(t)$ and $\mathbf{w}(t)$ are shown in Equation~\ref{eq:adLIF}, where $\alpha$, $\beta$, $a$ and $b$ are heterogeneously trainable parameters of the neuron. Finally, spikes $\mathbf{s}(t)$ are generated according to Equation~\ref{eq:adLif_spike}.

\begin{equation} \label{eq:adLIF_current}
    \mathbf{I}(t) = BNTT(W_f[x(t)]) + W_r[s(t-1)]
\end{equation}

\begin{equation}
    \label{eq:adLIF}
    \begin{aligned}
        \mathbf{u}(t) &= \alpha \left[ \mathbf{u}(t - 1) \right] + (1 - \alpha)\left[\mathbf{I}(t) - \mathbf{w}(t - 1)\right] - \theta [\mathbf{s}(t - 1)] \\
        \mathbf{w}(t) &= \beta [\mathbf{w}(t - 1)] + a (1 - \beta)  [\mathbf{u}(t - 1)] + b [\mathbf{s}(t - 1)] \\
    \end{aligned}
\end{equation}

\begin{equation} \label{eq:adLif_spike}
    \mathbf{s}(t) = 
    \begin{cases}
        0 \text{ if } \mathbf{u}(t) < \theta \\
        1 \text{ if } \mathbf{u}(t) \geq \theta
    \end{cases}
\end{equation}

The last layer of the network is a readout linear classifier, and the class corresponding to the maximum of the summation of output activities over all timesteps is chosen as the network prediction.
The RSNN network is trained with backpropagation through time using a boxed pseudo-gradient and cross-entropy loss.

\begin{center}
\begin{minipage}{.9\linewidth}
\begin{algorithm}[H]
\caption{Few-Shot Class-Incremental Learning with Prototypes}
\label{alg:prototypical_continual_learning}
\textbf{Requires:} Pre-trained network $g \circ f$ consisting of feature extractor $f$ and classifier $g: x 	\mapsto Wx+b$ \\
\textbf{Define:} $(x)_l$, $w_l$ and $b_l$ respectively the set of input samples, classifier weights and biases associated with a class $l$
\begin{algorithmic}[1]
\For{each base class $k$}
\State Compute prototype embedding $c_k = \text{Mean}[f((x)_k)]$ ~~~~~~~~~~~~~~~~ \textit{(also summed over time for SNN baseline)}
\State Compute corresponding classifier weights $w_k = 2c_k$ and biases $b_k = - c_kc_k^T$
\EndFor
\State Replace classifier layer $g$ with prototype weights: $W \leftarrow W_{B} = (w_k)_{k \in B}$ and biases $b \leftarrow b_{B} = (b_k)_{k \in B}$
\For{each session $i$ in sessions}
    \State Get session support $S^i$
    \State Repeat lines \textit{1} to \textit{4} for all new classes of $S^i$ to get prototype weights $W_{S^i}$ and biases $b_{S^i}$
    \State Extend the classifier layer weights $W \leftarrow [W, W_{S^i}]$ and $b \leftarrow [b, b_{S^i}]$
\EndFor
\end{algorithmic}
\end{algorithm}
\end{minipage}
\end{center}

We implement baseline solutions for the FSCIL task with both ANN and SNN models. The \textbf{frozen} baselines do not learn any new classes while the \textbf{prototypical} baselines follow the prototypical networks approach~\cite{snell2017prototypical} to classify new classes. For both baselines, the ANN and SNN models are pre-trained on the 100 base classes $B$, which employs the abundant number of samples to develop a robust feature extractor $f$, which generates embeddings from hidden layers that are passed to a readout classifier.

For the \textbf{frozen} baselines, the models parameters are frozen after pre-training for inference during all incremental sessions, thus setting a `worst-case' reference with no incremental learning but also no risk of catastrophic forgetting.

For the \textbf{prototypical} baselines, the pre-trained models learn 100 extra classes within the 10 incremental sessions in a 5-shot learning scenario. The prototypical networks protocol is applied in each incremental session as shown in Algorithm \ref{alg:prototypical_continual_learning}. 
Prototypical networks provide a clustering algorithm for classification that is equivalent to a readout affine operation on feature embeddings, resulting in a linear layer of weights and biases. Each class $k$ is represented by a prototype vector $c_k = \text{Mean}[f((x)_k)]$ defined as the average feature embedding produced by $f$ over all corresponding training samples $(x)_k$. The readout classifier layer is defined based on this prototype such that the weights $w_k$ and biases $b_k$ associated with class $k$ follow $w_k = 2c_k$ and $b_k = - c_kc_k^T$, which associates embeddings with the closest prototype with respect to the squared Euclidean distance~\cite{snell2017prototypical}.

For the SNN baselines, as the features also have a temporal dimensionality, we accumulate embeddings over all timesteps $t$ to define the prototype vector $c_k = \text{Mean}[\sum_t(f((x)_k)_t)]$. Also, as we maintain the summation over timesteps after the final prototype layer to keep the online nature of the SNN baseline, the biases will be applied at each timestep. Thus to maintain the balance between weighted inputs and biases, for the SNN baseline we also normalize the biases by the total number of timesteps $T$: $b_k = - c_kc_k^T/T$.

We fit the prototypical networks approach to the FSCIL task by first discarding the original output layer and replacing it with the prototype weights $W_{B}$ and biases $b_{B}$ of the base classes, computed as described above based on the averaged feature embeddings over all 500 training samples per base class. This causes an initial accuracy drop, as the trained output layer weights are replaced by clustered weights for the prototypical learning approach.
Then, for each incremental session, the prototype of each of the 10 new classes is defined based on the 5 corresponding support samples. The prototype weights and biases are computed in the same manner and concatenated to the existing classifier layer to accommodate for the new classes.

\subsubsection*{Event Camera Object Detection}
For both the RED ANN and Hybrid ANN-SNN baselines, the event data from the event camera are converted into frame-based representations using multi-channel time surfaces. Non-overlapping 50~ms time bins (with 50~ms stride), are further subdivided into three sub-bins. Each sub-bin, starting at timestamp $t_0$, generates two time surfaces $TS$ (Equation~\ref{eq:timesurfaces}), based on each event $(x, y, p, t)$ in the sub-bin, where $x, y$ are event coordinates, $p$ is positive or negative polarity, and $t$ is the event time.

\begin{equation}
\begin{aligned}
TS(p, y, x)=t-t_{0} \text { for each event }(x, y, p, t) \text { in the sub-bin.}
\end{aligned}
\label{eq:timesurfaces}
\end{equation}

The RED ANN \cite{Perot2020} is a deep convolutional neural network model using three feed-forward squeeze-and-excite~\cite{hu18squeeze} convolution layers followed by five recurrent convolution-LSTM~\cite{shi15convlstm} (ConvLSTM) layers. The squeeze-and-excite layers provide effective feature extraction while the ConvLSTM layers provide effective temporal learning.
The single-shot detection (SSD~\cite{Liu_2016}) head is used to predict the location and class of the bounding box based on multi-scale outputs from the recurrent layers. 
        
The Hybrid ANN-SNN architecture adopts five LIF spiking neural layers to replace the ConvLSTM layers in RED, and shares the same feed-forward convolutional blocks as the RED. The LIF neuron layers are connected with feed-forward convolution, and have far fewer weights than the ConvLSTM layers.
The Hybrid model uses the same input encoding method, object detection head, and training loss functions as the RED model. The LIF units are built using the SpikingJelly library~\cite{SpikingJelly}, and the neuron dynamics of the LIF membrane potential are given in Equations~\ref{eq:spikingjelly_h},~\ref{eq:spikingjelly_u}, and~\ref{eq:spikingjelly_spike}. $\mathbf{h}(t)$ is the charged potential before spiking during a timestep, dependent on activation input $X(t)$, and membrane time contant $\tau$, and $\mathbf{u}(t)$ is the final potential of the timestep which resets to the reset value $V_{reset}$ if $\mathbf{h}(t)$ reaches the threshold voltage $V_{th}$. The same thresholds determine $\mathbf{s}(t)$, whether a spike is produced. In the experiments, $\tau$ is set to 2.0; $V_{th}$ is 1.0, and $V_{reset}$ is 0.0.

\begin{equation}
    \mathbf{h}(t) = \mathbf{u}(t-1) + \frac{1}{\tau}(X(t) - \mathbf{u}(t-1))
    \label{eq:spikingjelly_h}
\end{equation}

\begin{equation}
    \mathbf{u}(t) = 
    \begin{cases}
        \mathbf{h}(t) & \text{if } \mathbf{h}(t) < V_{th} \\
        V_{reset} & \text{if } \mathbf{h}(t) \geq V_{th}
    \end{cases}
    \label{eq:spikingjelly_u}
\end{equation}

\begin{equation}
    \mathbf{s}(t) = 
    \begin{cases}
        0 & \text{if } \mathbf{h}(t) < V_{th} \\
        1 & \text{if } \mathbf{h}(t) \geq V_{th}
    \end{cases}
    \label{eq:spikingjelly_spike}
\end{equation}

The losses used to train the RED ANN and Hybrid baselines match previous work~\cite{Perot2020}, using a combination of regression and classification loss functions. Regression loss $L_r$ (Equation~\ref{eq:objdet_regression_loss}) for all predicted boxes $B$ and ground-truth boxes $T$ is given by smooth $l1$ loss $L_s$~\cite{Liu_2016} (Equation~\ref{eq:objdet_smoothl1}), averaged over $N$ predicted bounding boxes $B_i$ and their corresponding ground-truth boxes $T_i$. Smooth $l1$ loss is a piecewise loss function with threshold $\beta$, which is set to 0.11. For the classification loss ${L}_{c}$ (Equation~\ref{eq:objdet_classification_loss}), softmax focal loss~\cite{lin20focal} is used, with correct-class probability $p_l$ for all default boxes in the regression head and constant $\gamma$, which is set to 2.

\begin{equation}
\begin{aligned}
L_r(B, T)= \frac{1}{N} \sum_{j} L_s\left(B_{i}, T_{i}\right)
\end{aligned} \label{eq:objdet_regression_loss}
\end{equation}

\begin{equation}
{L}_{s}\left(B_{i}, T_{i}\right)=
\begin{cases}
\left|B_{i}-T_{i}\right|-\frac{\beta}{2} & \text { if }\left|B_{i}-T_{i}\right| \geq \beta \\
\frac{1}{2 \beta}\left(B_{i}-T_{i}\right)^{2} & \text { otherwise }
\end{cases} \label{eq:objdet_smoothl1}
\end{equation}

\begin{equation}
\begin{aligned}
L_c\left(p_{{l}}\right)=-\left(1-p_{{l}}\right)^{\gamma} \log \left(p_{{l}}\right)
\end{aligned} \label{eq:objdet_classification_loss}
\end{equation}


\subsubsection*{Non-human Primate Motor Prediction}






All baseline models have linear feed-forward layer architectures, where ANN, ANN\_Flat, and SNN\_Flat have topologies $N_{ch}-32-48-2$, and SNN uses $N_{ch}-50-2$. The varying topologies between SNN and SNN\_Flat attempt to optimize for complexity in the former and correctness in the latter.

The LIF neurons used in the SNN networks are developed using snnTorch~\cite{eshraghian2021training}, and have potential dynamics shown in Equations~\ref{eq:snntorch_u} and~\ref{eq:snntorch_spike}. Note that unlike the SpikingJelly neurons (Equations~\ref{eq:spikingjelly_h},~\ref{eq:spikingjelly_u}, and~\ref{eq:spikingjelly_spike}), the potential $\mathbf{u}(t)$ is reset in the timestep following a spike, rather than during the same timestep. As before, $V_{reset}$ is 0.0 and $V_{th}$ is 1.0, while $\beta$ is 0.96 for the SNN baseline and 0.50 for the SNN\_Flat baseline. The potential of the readout neurons in both baselines is directly read to produce velocity predictions, thus there is no spiking or reset mechanism and the neurons function as leaky accumulators.

\begin{equation}
    \mathbf{u}(t) = 
    \begin{cases}
        \beta \mathbf{u}(t-1) + X(t) & \text{if } \mathbf{s}(t-1) = 0 \\
        V_{reset} & \text{if } \mathbf{s}(t-1) = 1
    \end{cases}
    \label{eq:snntorch_u}
\end{equation}

\begin{equation}
    \mathbf{s}(t) = 
    \begin{cases}
        0 & \text{if } \mathbf{u}(t) \leq V_{th} \\
        1 & \text{if } \mathbf{u}(t) > V_{th}
    \end{cases}
    \label{eq:snntorch_spike}
\end{equation}

ANN, ANN\_Flat, and SNN\_Flat are trained using mean-squared error (MSE) loss over 50 epochs. The SNN baseline used a sliding window of 50 consecutive data points, representing 200~ms of data (50-point window, single-point stride) in order to calculate the loss, to allow for more information for backpropagation and avoid dead neurons and vanishing gradients. The MSE loss was linearly weighted from 0 to 1 for the 50 points within the window. The SNN was trained with 10-fold cross-validation, using an early-stopping regime with patience (epochs for which there is no improvement to the validation set) of 10 epochs.

\subsubsection*{Chaotic Function Prediction}

The LSTM baseline uses one LSTM layer followed by a ReLU activation and linear readout layer. As input, the LSTM uses an explicit memory buffer of the last $M=50$ points. During training, input $\mathbf{x}(t)$ to the LSTM uses the Mackey-Glass data $\mathbf{f}(t)$ (Equation~\ref{eq:lstm_training}), whereas, during autoregressive evaluation, the input uses prior predictions $\mathbf{y}(t)$ (Equation~\ref{eq:lstm_autoregressive}). Values $\mathbf{u}(t<0)$ and $\mathbf{v}(t<0)$ are zero.

\begin{equation} \label{eq:lstm_training}
    \mathbf{x}(t) = (\mathbf{f}(t-M), \mathbf{f}(t-M-1), \ldots, \mathbf{f}(t))
\end{equation} 

\begin{equation} \label{eq:lstm_autoregressive}
    \mathbf{x}(t) = (\mathbf{y}(t-M-1), \mathbf{y}(t-M-2), \ldots, \mathbf{y}(t-1))
\end{equation}

The LSTM is trained using MSE loss for backpropagation with 200 epochs. 
The hyperparameter sweep used the evaluation setup of 30 instantiations of $\tau = 17$ Mackey-Glass data, with each instance shifted forward by half of the Lyapunov time. The corresponding sets with the lowest sMAPE scores were used to report the results.

For the ESN, the standard architecture with one hidden layer (i.e., reservoir) with recurrent connections was used, where the states of the reservoir $\mathbf{r}(t) \in \mathbb{R}^{D}$ at timesteps $t$ are evolving according to the dynamics shown in Equation~\ref{eq:esnres}. 
The random matrix $\mathbf{W}^{\mathrm{in}} \in \mathbb{R}^{D  \times d+1}$ with components drawn from the uniform distribution projects $d$-dimensional input $\mathbf{f}(t)$ ($d=1$ for the Mackey-Glass system), augmented with constant bias, into $D$ neurons of the reservoir. The recurrent connectivity is defined by the second (potentially sparse) random matrix $\mathbf{W} \in \mathbb{R}^{D  \times D}$ 
with nonzero components drawn from the normal distribution; 
$\alpha$, $\gamma$, and $\beta$ are hyperparameters controlling the behavior of the ESN.

\begin{equation}
\mathbf{r}(t)=(1-\alpha)\mathbf{r}(t-1)+\alpha \tanh \left(\gamma\mathbf{W}\mathbf{r}(t-1) + \beta\mathbf{W}^{\mathrm{in}} [1; \mathbf{f}(t)] \right)
\label{eq:esnres}
 \end{equation}


To make a prediction $\mathbf{y}(t)$, the ESN uses the readout matrix $\mathbf{W}^{\mathrm{out}} \in \mathbb{R}^{d  \times D+d+1}$ that computes the activation of the output layer based on the current states of the input and hidden layers: 
$\mathbf{y}(t) = \mathbf{W}^{\mathrm{out}} [\mathbf{f}(t); \mathbf{r}(t)]$. 
To predict the values of the system at the next timestep, i.e. $\mathbf{y}(t)$ predicts $\mathbf{f}(t+1)$, the output layer has $d$ neurons. 

The training of $\mathbf{W}^{\mathrm{out}}$ is formulated as a linear regression problem so that it can be computed with the regularized least squares estimator (Equation~\ref{eq:readout:regr}), where $\mathbf{H} \in \mathbb{R}^{M \times D+d+1}$ is an activation matrix that stores the readout for $M$ timesteps in the training data, $\mathbf{Y} \in \mathbb{R}^{M \times d}$ is another matrix that stores the corresponding ground-truth values for the same timesteps, and $\lambda$ is the regularization parameter of the estimator.

\begin{equation}
\mathbf{W}^{\mathrm{out}}= \mathbf{Y}^{\top} \mathbf{H} \left( \mathbf{H}^{\top} \mathbf{H} + \lambda \mathbf{I}\right)^{-1}
\label{eq:readout:regr}
\end{equation}

Like for the LSTM, optimal hyperparameters are chosen based on lowest average sMAPE score over 30 time series. For each series, the ESN weight matrices $\mathbf{W}^{\mathrm{in}}$ and  $\mathbf{W}$ were randomly initialized. The corresponding sets with the lowest sMAPE scores were used to report the results.

\subsection*{System Track Metrics}
\secondrev{Given the variability of task application areas and system sizes, the NeuroBench metric methodology is defined on a per-benchmark basis.} \secondrev{Particularly for efficiency metrics, as neuromorphic systems are not matured, a singular power measurement method cannot be applied to all submissions. It is the responsibility of the submitter to faithfully capture all active processing components for their system and fairly outline their methodology in the report. As NeuroBench moves towards head-to-head benchmarking of neuromorphic systems by hardware vendors and owners, official results must be associated with a report that provides context for the benchmark submission results, as the overall benchmark format is generally open and does not have stringent consistency rules. The report must contain 
\begin{itemize}
    \item an outline of the system architecture,
    \item an outline of the algorithm used (model architecture, tuning),
    \item a diagram depicting the workflow, including (where applicable) data initialization, host pre-processing, data loading, on-device preprocessing, inference, post-processing,
    \item timing measurement description,
    \item power measurement description, including measurement devices, included hardware components, and measurement time resolution,
    \item a re-iteration of the results.
\end{itemize}}

\secondrev{In addition, official submissions will be subject to potential audits during which an auditor will inspect the methodology and request additions or revisions to the results and report if necessary.}



\subsection*{System Track Benchmarks}
\secondrev{Detailed information on each v1.0 system track benchmark is provided here, and the most updated information can be found in the official NeuroBench system track documentation: \url{https://github.com/NeuroBench/system_benchmarks}}

\subsubsection*{Acoustic Scene Classification (ASC)}
\secondrev{\paragraph{Benchmark Dataset} The dataset is based on the DCASE 2020 acoustic scene classification challenge~\cite{Heittola2020}, using the TAU Urban Acoustic Scenes 2020 Mobile datasets.
Of the 10 available scene classes, 4 are used: ``airport'', ``street\_traffic'', ``bus'', ``park''.
Of the 9 real / simulated audio recoding devices available, 1 real device is used: ``a''.
Audio samples are sliced into 1-second samples. The audio may be resampled to a different frequency as a pre-processing step, which is not included in inference measurement.
41360 training samples are available, as well as 16240 test samples.
The NeuroBench system track repository linked above provides a download script and a PyTorch-compatible dataset file which is expected to be used as the front-end data generator for all submissions. The data may be reformatted into a different framework, this is not included in inference measurement.}

\secondrev{\paragraph{Task} After training a model using the training set, the submitter will report test set accuracy, as well as execution time and energy per inference on the system. The test split should not be used to train or tune the model, only the train split.
Audio samples will be processed in batch-size 1, in which one sample is processed at a time and the next sample is not processed until the previous one is finished.
The general compute flow consists of three steps: pre-processing, inference, and classification. 
One inference is defined as the processing of one second of audio data. Inference is separate from classification because systems are often intended to process samples as sequences over time, rather than all at once, and the classification may be available before the whole data sequence is seen. Classification thus does not need to be included in the benchmark measurement.
Pre-processing in general can be defined as feature extraction or the conversion of raw audio data into a format that is suitable for inference. Execution time and energy of pre-processing must be included in benchmark measurement, and will be reported separately from inference. 
In certain systems, e.g. Synsense Xylo, pre-processing blocks use analog hardware on-chip to directly convert real-time analog microphone output to digital spikes for inference. In order to operate inference from a digitally-encoded dataset, this pre-processing must be simulated and the spikes are sent through a side-channel directly to the inference block. For such cases, the reported pre-processing measurements will be for the analog hardware running in real-time on the dataset audio played over speakers into the device microphone.
}

\secondrev{\paragraph{Metrics} The following metrics should be reported: \begin{itemize}
    \item \textbf{Accuracy} -- Accuracy of the predictions on the test set, measured from the system and not in any software simulation.
    \item \textbf{Execution time} -- Execution time is the average time per sample for pre-processing and inference. The final result should be averaged over all samples in the test set. The time begins when data has been loaded into on-board or on-chip memory and is prepared to be processed. The time ends when the last timestep of the sequence has completed processing.
    \item \textbf{Power/Energy} -- Idle power and active power should be reported, where idle power is power of the chip when it has been configured and it has prepared to begin processing. Note that this should not measure the device in a lower-power sleeping state. Active power is the average power consumed during processing. The difference between active power and idle power should be reported as dynamic power. Power should be converted to energy by multiplying power by the averaged execution time. The power measurement should include all computational components and power domains which are active during the workload processing. If applicable, power may be measured over longer data sequences (e.g. 5 seconds rather than 1 second), such that configuration costs are amortized over a longer period of processing. This should be done separately from accuracy and execution time experiments, and should be clearly detailed in the associated submission report.
\end{itemize}}

\subsubsection*{Quadratic Unconstrained Binary Optimization (QUBO)}

\secondrev{Quadratic Unconstrained Binary Optimization (QUBO) refers to the problem of finding the binary variable assignment $x_i\in\{0, 1\}$ that optimizes the quadratic cost function
$$ \min_{\mathbf{x}\in\{0,1\}^n} c(\mathbf{x}) = \min_{\mathbf{x}\in\{0,1\}^n} \mathbf{x}^T\mathbf{Q}\mathbf{x}$$
subject to no constraints.}

\secondrev{The solvers for QUBO will be benchmarked using Maximum Independent Set (MIS) workloads. Given an undirected graph $\mathcal{G}=(\mathcal{V}, \mathcal{E})$, an independent set $\mathcal{I}$ is a subset of $\mathcal{V}$ such that, for any two vertices $u, v \in \mathcal{I}$, there is no edge connecting them, i.e., $\nexists \; e \in \mathcal{E} \;s.t.\; e=(u,v) \;\vee\; e=(v,u)$.}

\secondrev{The MIS problem has a natural QUBO formulation: for each node $u\in\mathcal{V}$ in the graph, a binary variable $x_u$ is introduced to model the inclusion or not of $u$ in the candidate solution. Summing the quadratic terms $x_u^2$ will thus result in the size of the set of selected nodes. To penalize the selection of nodes that are not mutually independent, a penalization term is associated to the interactions $x_ux_v$ if $u$ and $v$ are connected. The resulting $\mathbf{Q}$ matrix coefficients are defined as
$$
q_{uv} = \begin{cases}
    -1 &\text{if } u = v \\
    4 &\text{if } u \neq v \text{ and } (u, v) \in \mathcal{E} \\
    0 & \text{otherwise.}
\end{cases}
$$
The MIS problem is NP-hard and intractable, and therefore any solver system approximates a solution. Therefore, the cost of the QUBO formulation ($\mathbf{x}^T\mathbf{Q}\mathbf{x}$) is used to assess solutions, and solutions are not restricted to being maximum nor independent sets. The QUBO formulation ensures that any non-independent set will always have a higher cost than a corresponding independent set with conflicting nodes removed.}

\secondrev{\paragraph{Benchmark Dataset} The benchmark's workload complexity is defined such that it can automatically grow over time, as neuromorphic systems mature and are able to support larger problems.
\begin{itemize}
    \item Number of nodes, spaced in a pseudo-geometric progression: 10, 25, 50, 100, 250, 500, 1000, 2500, 5000, ...
    \item Density of edges: 1\%, 5\%, 10\%, 25\%.
    \item Problem seeds: 0, 1, 2, 3, 4 are allowed for tuning. At evaluation time, for official results NeuroBench will announce five seeds for submission. Unofficial results may use seeds which are randomly generated at runtime.
\end{itemize}
}
\secondrev{Each workload will be associated with a target optimality, which is the minimum cost found using a conventional solver algorithm. Small QUBO workloads with fewer than 50 nodes will be solved to global optimality, corresponding to the true maximum independent set. Larger workloads cannot be reasonably globally solved. The DWave Tabu CPU sampler~\cite{dwavesamplers} will be used with 100 reads and 50 restarts, and the QUBO solution with the best cost (best-known solution, BKS) found will set the target optimality for the tuning workload seeds. For evaluation workload seeds, the same method will be used to set the target optimality. 
NeuroBench will provide target optimalities for workloads up to 5000 nodes. Submissions are encouraged to continue scaling up the workload size along the pattern to demonstrate the capacity of their systems. The first group that tackles workloads of an unprecedented size should provide the benchmark solutions via a pull request to the system track repository. The dataset workload generator and scripts for the DWave Tabu sampler to compute optimal costs are available in the repository, and this code is expected to be used as the front-end data generator for all submissions.
}

\secondrev{\paragraph{Task and Metrics} Based on the BKS found for each workload, the BKS-Gap optimality score of the solution found by the SUT is defined as
$$
\text{BKS-Gap} = (\frac{c - c_{\mathrm{target}}} {c_\mathrm{target}})\ ,
$$ where $c_\mathrm{target}$ is the QUBO cost of the BKS, and $c$ is the cost found by the SUT. This may be reported as a percentage gap by multiplying by 100. If the SUT manages to beat the BKS, then the BKS-Gap will be negative.}

\secondrev{Given each workload, the benchmark should report the BKS-Gap of the solution found by the SUT after a fixed runtime. The time begins after the graph has been loaded into the SUT. Timeouts spread across orders of magnitude ($10^{-3}$, $10^{-2}$, $10^{-1}$, $10^{0}$, $10^{1}$, $10^{2}$). As the runtime is fixed, no measured timing metric is reported, and submissions should report average power over the duration of the runtime, as it is directly proportional to energy consumed. Importantly, the QUBO solver needs a module to measure the cost of its solutions, and this module should be considered as part of the SUT, thus its computational demand and power must be included in the benchmarking results.}

\secondrev{In the future, a different optimization task scenario may be used for the same QUBO dataset, in which the SUT must run until it reaches a small BKS-Gap, rather than stopping after a fixed-timeout. Here, the benchmark should report latency and energy required to reach BKS-Gap thresholds, e.g., 0.1, 0.05, and 0.01. This task scenario normalizes systems to solution quality, rather than SUT runtime.}

\subsection*{System Track Baselines}

\subsubsection*{Acoustic Scene Classification (ASC)}
\begin{figure}[t]
    \centering
    \begin{subfigure}{0.48\textwidth}
        \includegraphics[width=\linewidth]{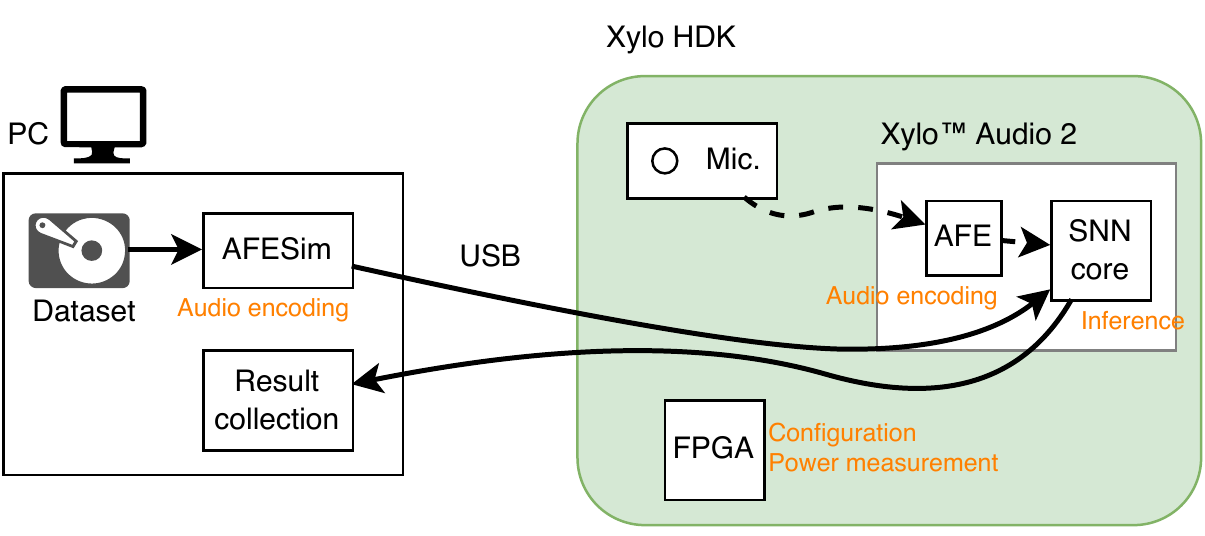}
    \vspace*{1cm}
    \end{subfigure}
    \hspace*{0.5cm}
    \begin{subfigure}{0.3\textwidth}
        \includegraphics[width=\linewidth]{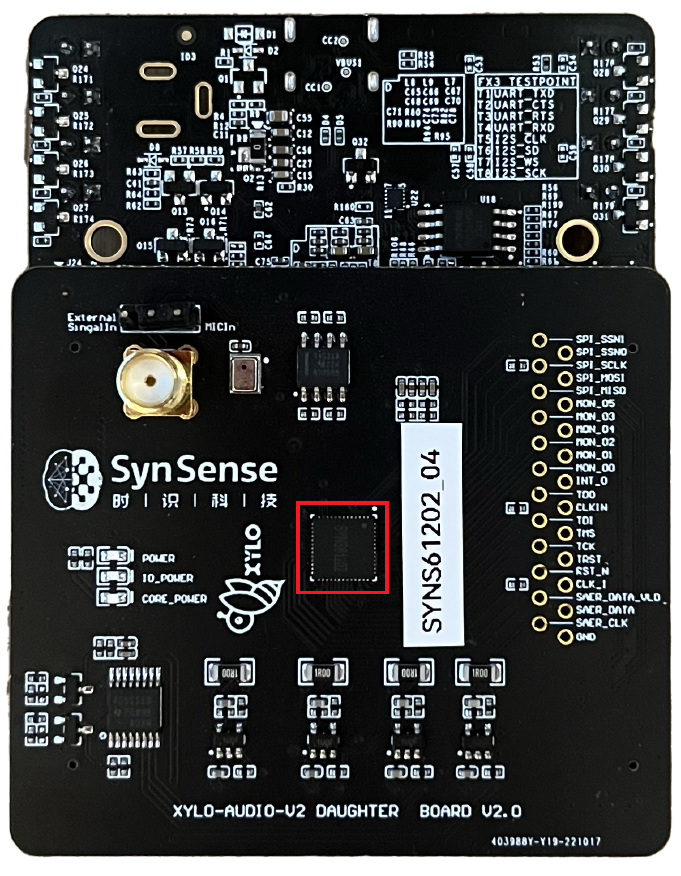}
    \end{subfigure}
        \caption{\secondrev{Left: An overview of the benchmarking system for the Xylo. A simulator of the analog encoding module is run on a PC and streamed to the SNN inference core via USB. After inference, outputs are routed back to the PC for classification. An on-board FPGA configures and records power of the Xylo components. Right: The Xylo\texttrademark Audio 2 hardware development kit (HDK), used as the SUT. The red outline marks the Xylo inference module. Figures taken, with permission, from Ke et al.~\cite{ke2024neurobenchdcase2020acoustic}}}
        \label{fig:xylo}
\end{figure}

\secondrev{\paragraph{CPU} The CPU baseline uses an Arduino Nano 33 BLE board, which runs preprocessing and inference on an ARM Cortex-M4F microcontroller running at 64MHz. All training and deployment uses Edge Impulse, a commercially developed ML-operations platform for tinyML~\cite{hymel2023edgeimpulsemlopsplatform}. The 1 second digital audio samples were converted into two-dimensional frames of Mel-filterbank energies (MFE), and inference uses a network with two convolution layers with batch normalization and max pooling. The trained model was quantized then compiled to an Arduino library using Edge Impulse EON compiler, which optimizes for memory and flash usage. Execution time is measured using on-chip timers from the Arduino, while power is measured using an external multimeter for total system power. Power was measured separately for idle, active pre-processing, and active inference by taking the average power over 60 seconds. The Arduino repeatedly computes over one sample loaded in memory, with a recording frequency of 3 Hz on a Keysight 34465A digital multimeter.}

\secondrev{\paragraph{Synsense Xylo} As shown in Figure~\ref{fig:xylo}, the Xylo baseline used a host PC to run a simulation of the analog pre-processing unit on the benchmark dataset, which was routed into the SNN core on the Xylo SUT. Accuracy is measured by performing a maximum on the output spikes of the SNN. The SNN on Xylo is a feedforward network with three hidden layers, where each layer is connected by synapses with varying time delays. Training is done using the open-source Rockpool toolchain~\cite{rockpool2019}. In order to amortize measurement overheads, power and execution time are measured over continuous streams of 10 seconds of audio, where the execution time result is divided by 10 and the power result is averaged over the duration. The measurements are made by the on-board FPGA, which operates at 12.5MHz and samples power at 1280 Hz. Accuracy is still measured on the 1-second samples. Full details are available in Ke et al.~\cite{ke2024neurobenchdcase2020acoustic}}

\subsubsection*{Quadratic Unconstrained Binary Optimization (QUBO)}
\begin{figure}[t]
    \centering
        \includegraphics[width=0.85\linewidth]{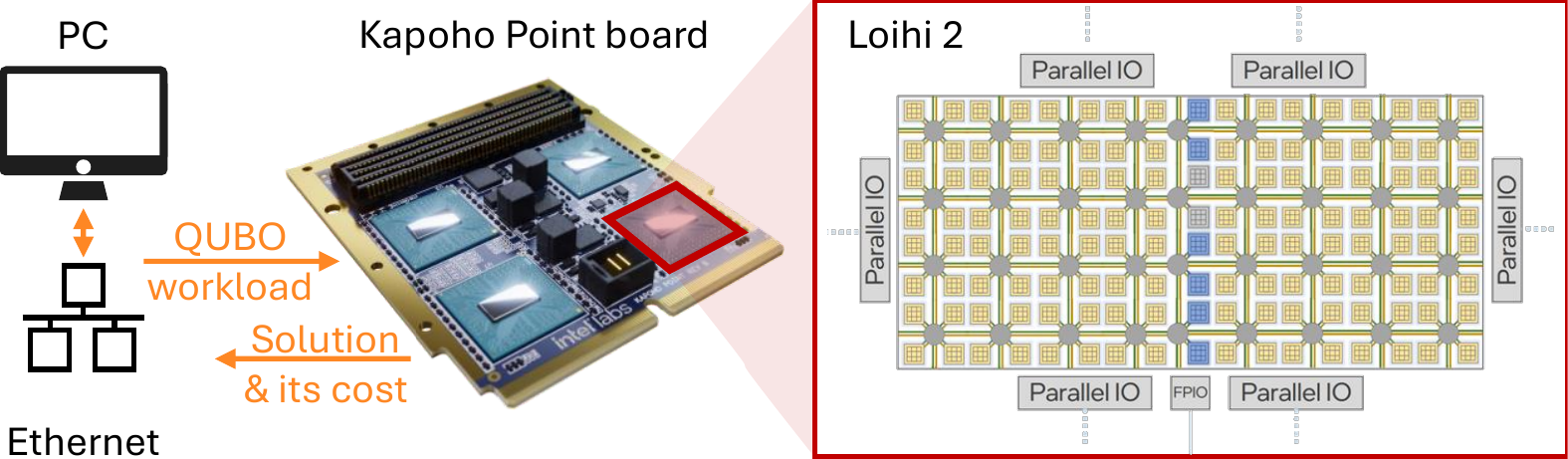}
        \caption{\secondrev{A Kapoho Point board with Loihi 2 chips is connected to a host PC via ethernet. Shown here is a Kapoho Point board with 8 Loihi 2 chips, while the experiments were run on a board with 1 Loihi 2 chip. The QUBO workload is loaded onto the chip, and all processing happens on Loihi 2. On Loihi 2, the neuro-cores, in yellow, are arranged in two $8\times{}8$ grids and iteratively update the variables of the QUBO workload. Embedded CPUs, shown in blue, monitor the cost of the variable assignment. Six parallel IO ports enable 3D stacking of multiple chips for larger workloads than used here. An additional IO port provides the communication to the host PC. The energy and runtime measurements cover the computation of the Loihi 2 chip.}}
        \label{fig:loihi2}
\end{figure}

\secondrev{The processors ran repeatedly with five different initial variable assignments and seeds for different timeouts between $10^{-3}$ s to $10^{3}$ s, and for different workloads sizes of up to $1000$ variables. The neuromorphic algorithm is a parallelized version of simulated annealing that was running on a Kapoho Point board with one Loihi 2 chip, as shown in Figure~\ref{fig:loihi2}. The Loihi 2 board was controlled using Lava \texttt{0.8.0} and Lava Optimization \texttt{0.3.0}. For comparison with conventional hardware, two solvers were adopted based on simulated annealing and tabu search, as implemented in the D-Wave Samplers \texttt{v1.1.0} library \cite{dwavesamplers}. The library was compiled on Ubuntu \texttt{20.04.6 LTS} with GCC \texttt{9.4.0}  and Python \texttt{3.8.10}. CPU measurements were obtained on a machine with Intel Core i9-7920X CPU @ 2.90 GHz and 128GB of DDR4 RAM, using Intel SoC Watch for Linux OS \texttt{2023.2.0}. All details on the solvers, benchmarking routine, and results have been provided by Pierro et al.~\cite{pierro2024solvingquboloihi2}.
}

\section*{Acknowledgements}

Authors of this work have been supported in parts by Semiconductor Research Corporation (JY), the European Research Council (ERC) under the European Union’s Horizon 2020 research and innovation programme (grant agreement No.~101001448), a grant from the Research Grants Council of the Hong Kong Special Administrative Region, China [Project No.~CityU 11200922], ARC Laureate Fellowship FL210100156,  and the EU H2020 project BeFerroSynaptic (871737). We acknowledge the financial support of the CogniGron research center and the Ubbo Emmius Funds (Univ. of Groningen). We acknowledge a contribution from the Italian National Recovery and Resilience Plan (NRRP), M4C2, funded by the European Union –NextGenerationEU (Project IR0000011, CUP B51E22000150006, “EBRAINS-Italy”). The work of SynSense was partially supported by the European Commission, under the Horizon grant Ferro4Edge AI (grant agreement 101135656). This work is partly funded by the German Federal Ministry of Education and Research (BMBF) and the free state of Saxony within the ScaDS.AI center of excellence for AI research and by the German Federal Ministry for Economic Affairs and Climate Action (BMWK) under contract 01MN23004F (ESCADE).

Sandia National Laboratories is a multi-mission laboratory managed and operated by National Technology \& Engineering Solutions of Sandia, LLC (NTESS), a wholly owned subsidiary of Honeywell International Inc., for the U.S. Department of Energy’s National Nuclear Security Administration (DOE/NNSA) under contract DE-NA0003525. This written work is authored by an employee of NTESS. The employee, not NTESS, owns the right, title and interest in and to the written work and is responsible for its contents. Any subjective views or opinions that might be expressed in the written work do not necessarily represent the views of the U.S. Government. The publisher acknowledges that the U.S. Government retains a non-exclusive, paid-up, irrevocable, world-wide license to publish or reproduce the published form of this written work or allow others to do so, for U.S. Government purposes. The DOE will provide public access to results of federally sponsored research in accordance with the DOE Public Access Plan. This paper describes objective technical results and analysis. Any subjective views or opinions that might be expressed in the paper do not necessarily represent the views of the U.S. Department of Energy or the United States Government.

\section*{Author contributions statement}
Authors are grouped based on contributions, and ordered alphabetically within groups.
JY led project discussions and management. JY and KVdB implemented harness metric infrastructure, conducted experiments, and prepared the manuscript.
The following authors primarily developed the main algorithm track results: DdB and MF on the keyword few-shot class-incremental learning task; GT and SW on the event camera object detection task; PH, PVS and BZ on the non-human primate motor prediction task; YB and DK on the chaotic function prediction task; and NP led harness infrastructure development. The following authors primarily developed the main system track results: WK and MAK on the acoustic scene classification task; AP and PS on the QUBO task.
SHA, GVJ, BL, AM, AKM, GL, and TS developed components of the algorithm track harness infrastructure.
ZA, MA, BA, AGA, CB, AB, PB, S. Bohte, S. Buckley, GC, EC, FC, GdC, A. Danielescu, A. Daram, MD, YD, JE, TF, JF, VF, SF, PMF, WG, AG, HAG, GI, SJ, VK, L. Khacef, JCK, L. Kriener, RK, DK, SL, YL, HM, RM, JMM, CM, KM, DRM, EN, TN, FO, AO, PP, JP, MP, C. Pehle, MAP, C. Posch, AR, YS, CJSS, AvS, J. Schemmel, S. Schmidgall, CS, J. Seo, S. Sheik, SBS, MS, AS, KS, MS, TCS, JT, NT, GU, MV, CMV, BV, AY, and FTZ participated in discussions during meetings, prepared sections for the present manuscript and/or its preprint, and reviewed the manuscript.
CF and VJR jointly supervised the project, reviewed the manuscript, and analyzed results.

\bibliography{ref}


\section*{Additional information}

\subsection*{Competing interests statement}

The NeuroBench benchmark framework was developed collaboratively and specifically to allow for as objective and applicable comparison as possible. The selection of initial benchmark tasks reflect the authors' research interests, which includes commercial interests for companies. These do not affect our results in any way, nor the value of our contributions. The benchmark harness and framework are open-source and intended to be further extended by the community over time.



\end{document}